\documentclass[runningheads]{llncs}

\usepackage[T1]{fontenc}
\usepackage{graphicx}
\usepackage{booktabs}
\usepackage{adjustbox}
\usepackage{amsmath}
\usepackage{xcolor}
\usepackage{tikz}
\usetikzlibrary{arrows.meta,positioning,calc}
\usepackage[section]{placeins}
\usepackage[colorlinks,allcolors=blue]{hyperref}
\hypersetup{pdfauthor={Choulhyouc Lee},pdftitle={Diagnosing and Correcting Skin-Tone Color Casts in Ultra-Lightweight Photo Retouching Networks},pdfkeywords={}}

\begin{document}

\title{Diagnosing and Correcting Skin-Tone Color Casts in\\ Ultra-Lightweight Photo Retouching Networks}
\author{Choulhyouc Lee}
\institute{SNOW Corporation, Republic of Korea\\ \email{iron.lee@snowcorp.com}}
\titlerunning{Correcting Skin-Tone Casts in Ultra-Lightweight Retouching}
\authorrunning{C. Lee}

\maketitle
\let\thefootnote\relax\footnotetext{Accepted to the Asian Conference on Computer Vision (ACCV) 2026. This is the author's version of the work, with the supplementary material appended; the final version will appear in the ACCV 2026 proceedings.}\addtocounter{footnote}{0}

\begin{abstract}
Color casts are expected to come from the scene or camera, and a retouching model removes them.
We report a case where the model leaves one: an ultra-lightweight retouching network renders
human skin greener than the original because it under-restores the skin chroma that a degradation
removes and, even on clean inputs, adds a small hue shift of its own, a behavior that recurs across architectures and, in a
matched FiveK control, tracks the training recipe. A
common paired-data-free recipe, synthetic self-supervision (degrade high-quality images, learn to
restore them), silently produces it while whole-frame PSNR/SSIM stay essentially unchanged. In a
representative 6{,}825-parameter histogram-conditioned network, the cast appears in 52 of 70
portraits under a ground-truth-anchored, white-balance-free protocol. Swapping the luminance source
does not remove it, so the artifact is chrominance-intrinsic. In a two-architecture, two-recipe comparison, both a histogram-conditioned and a CNN-conditioned
network develop the cast under synthetic self-supervision, yet both are near-neutral in-domain when
trained on MIT-Adobe FiveK expert pairs. Image-level objectives miss it, since global affine
models satisfy image-mean color losses by offsetting across regions, and a global-mean green loss
worsens the cast (52/70 to 57/70). With zero added parameters, our per-pixel, skin-masked, asymmetric green-excess loss
reduces the cast to 1/70 across stratified skin tones ($68.1\%$ to
$2.2\%$ over the full 70{,}000-image FFHQ corpus; $64.4\%$ to $0.8\%$ on held-out CelebA-HQ real
photos). In-domain FiveK PSNR/SSIM are unchanged ($+0.13$\,dB, $+0.004$), at the cost of a small
whole-frame CIEDE2000 increase ($8.77\!\to\!9.69$) and a magenta over-correction where skin is
nearly neutral (skin $\Delta a^*$ $-1.5\!\to\!+4.4$ and CIEDE2000 $5.6\!\to\!6.4$ on clean inputs; $\Delta a^*$ $+4.48$ when trained on real pairs): the loss targets the signed
green--magenta axis, not total skin-color fidelity. The same loss suppresses the
cast across three ultra-light architectures with distinct conditioning mechanisms. A whole-frame
green metric correlates with skin-region CIEDE2000 at the method level (Spearman $\rho = 0.89$),
with a moderate per-image correlation. The final network exports to a 35\,KB TFLite model.
\end{abstract}

\section{Introduction}\label{sec:intro}
\begin{figure}[t]\centering
\includegraphics[width=.82\linewidth]{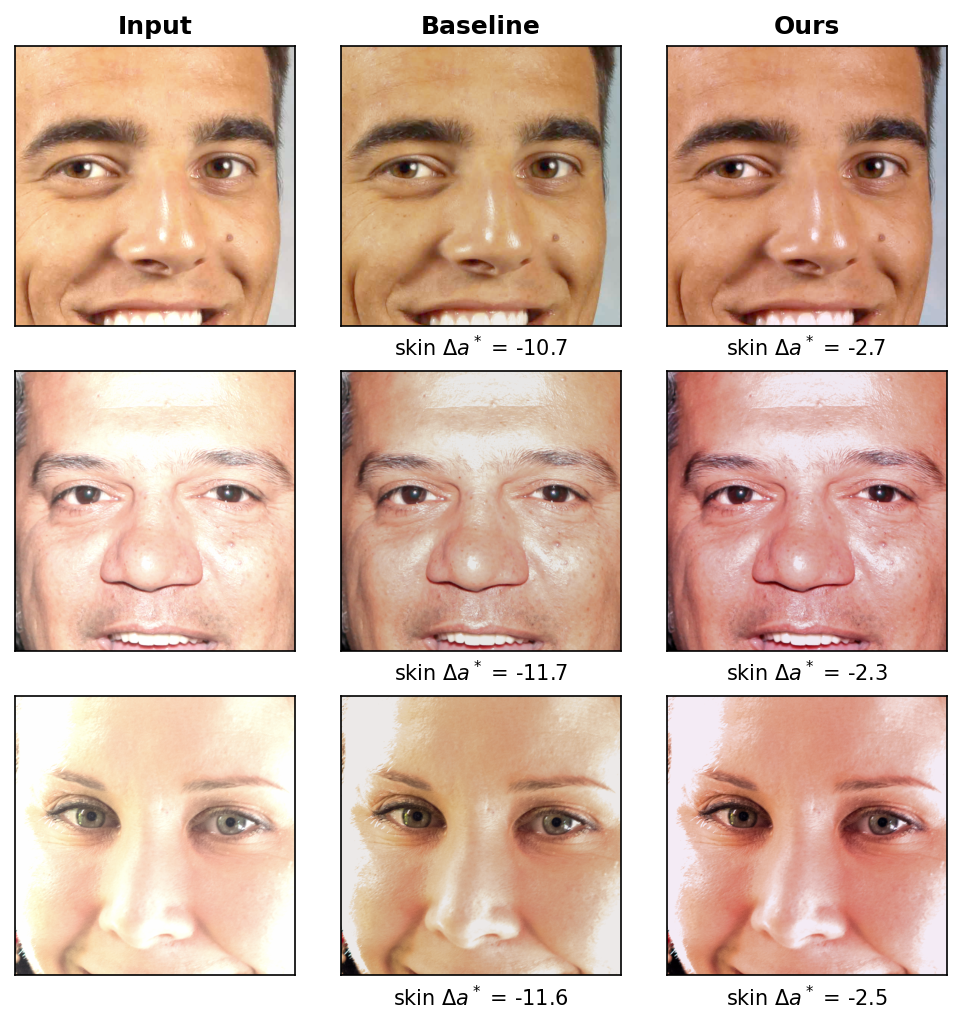}
\caption{Representative examples spanning the skin-tone strata of our protocol (dark/medium/light
by ground-truth skin luma, Sec.~\ref{sec:protocol}); columns: white-balance-free degraded input, the baseline
(v1g), ours. The synthetically trained baseline leaves skin visibly greener than the ground truth
(not shown); our per-pixel skin-masked loss substantially suppresses the cast at identical model size. Numbers report mean
skin $\Delta a^*$ inside the GT-derived skin mask ($<0$ = green, $>0$ = magenta). FFHQ photo credits:
supplementary material.}
\label{fig:teaser}
\end{figure}
Photo retouching, which restores pleasing color, exposure, and contrast to a degraded
photograph, increasingly runs on the device that captured the image~\cite{quant2018}. At this end of the
efficiency spectrum, \emph{ultra-lightweight} retouching networks with fewer than 40K
parameters~\cite{elirnet2025,csrnet2020,neurop2022} modulate the image with a handful of
global operators predicted from low-cost image statistics, and run in milliseconds on mobile
hardware. Because paired expert-retouched data is scarce and expensive, a common
paired-data-free training recipe at this scale is \emph{synthetic self-supervision}: high-quality
images are degraded with randomized brightness, contrast, saturation, gamma, and white-balance
perturbations, and the network is trained to undo the degradation. The recipe is attractive
because it needs no annotation and scales to any image collection.

This paper documents a hidden cost of that recipe. Training a recent 6.8K-parameter
histogram-conditioned retouching network (eLIR-Net~\cite{elirnet2025}; our channel-aware
\emph{v1g} variant, Sec.~\ref{sec:diag}) under the common
synthetic protocol, we find that the model \emph{leaves human skin greener than the original}
(Fig.~\ref{fig:teaser}): on a GT-anchored 70-portrait protocol, 52 of 70 restored portraits carry a measurable
green shift on skin (mean skin $\Delta a^* \approx -7$; the trend holds at scale, with $68.1\%$ of
all 70{,}000 FFHQ images flagged in a distribution audit, and transfers to held-out real photos,
Sec.~\ref{sec:exp}), because the degradation lowers skin
chroma and the model under-restores it; on clean inputs it adds a smaller shift of its own
(39/70 flagged). The defect is not captured by the commonly reported metrics: whole-frame
PSNR and SSIM are essentially unchanged, while signed skin $\Delta a^*$ exposes a systematic
green shift. Skin is a perceptually critical image region
with fairness implications, so a systematic skin cast that passes standard evaluation
can reach deployment undetected.

This paper makes four contributions.

\noindent\textbf{(i) Diagnosis of its nature.} The cast is \emph{chrominance-intrinsic}:
swapping the luminance source of the output (including the original network's own
luminance) leaves the green count unchanged (29/40 in a Y-source sweep). The cast is confined
to the chrominance channels, which explains why the natural two-model
luminance/chrominance ensemble fails to remove it in our setting (Sec.~\ref{sec:diag}). It
reflects under-restored chroma and a small hue shift, not a skin-specific chroma gain (Sec.~\ref{sec:ysweep}).

\noindent\textbf{(ii) Why image-mean objectives fail, and a corrective loss.} A global
affine model can satisfy any \emph{image-mean} color objective by regional offsetting
(over-correcting the background to cancel skin error in the mean), so a global-mean green loss
worsens the cast (52$\to$57/70). Our per-pixel, skin-masked, asymmetric
green-excess loss provides dense gradients inside the skin region and, being one-sided,
penalizes only green in excess of the ground truth. It reduces the cast to
\textbf{1/70} across stratified dark/medium/light skin tones (5$\to$0, 22$\to$0, 25$\to$1) at
zero parameter cost with in-domain FiveK PSNR/SSIM unchanged; the same loss suppresses
the cast across three ultra-light architectures with distinct conditioning mechanisms
(Sec.~\ref{sec:method},~\ref{sec:exp}).

\noindent\textbf{(iii) Localizing the cause.} In an observational two-architecture, two-recipe
comparison, both the eLIR architecture and
CSRNet~\cite{csrnet2020} (CNN-conditioned) develop the cast under synthetic self-supervision, yet
both are near-neutral in-domain when trained on real expert pairs (MIT-Adobe
FiveK~\cite{fivek2011}; eLIR skin $\Delta a^*$ $-0.96$ vs.\ $-8.3$ under the synthetic recipe).
A same-corpus FiveK control isolates the training-regime effect for eLIR, and our loss suppresses the cast while preserving in-domain PSNR/SSIM
(Sec.~\ref{sec:exp}).

\noindent\textbf{(iv) Measurement.} Auditing a low-cost whole-frame green metric against
perceptual skin-region CIEDE2000 yields a method-level rank correlation of $0.89$ (per-image
$0.50$): the metric is reliable for
\emph{detecting} green but blind to magenta over-correction, so we report it paired with signed
skin $\Delta a^*$ (Sec.~\ref{sec:exp}). A residual brightness under-correction, which loss
reweighting did not move, is handled by a manual inference-time luma gain ($+5.0$\,dB on
over-exposed inputs, harmful on clean ones; Sec.~\ref{sec:bright}).

The final model is a single 6{,}825-parameter network that exports to a 35\,KB TFLite matching
the reference implementation to $2.4{\times}10^{-7}$ maximum error; code will be released.

\section{Related Work}\label{sec:related}

\paragraph{Efficient global retouching.} Learned enhancement spans spatial architectures such as
HDRNet's bilateral-grid affine coefficients~\cite{hdrnet}, encoder--decoder enhancers,
image-adaptive 3D LUTs~\cite{lut3d2020}, and local parametric filters~\cite{deeplpf2020}. At the extreme-efficiency end, \emph{global
operator} models predict a small set of image-level parameters: CSRNet's
condition-network-modulated pointwise convolutions~\cite{csrnet2020}, sequential neural color
operators~\cite{neurop2022}, curve-based methods such as Zero-DCE~\cite{zerodce}, and
eLIR-Net's histogram-conditioned global feature modulation at 6.8K
parameters~\cite{elirnet2025}, our subject model. These works optimize whole-frame fidelity and report whole-frame metrics;
region-specific chromatic side effects of the \emph{training recipe}, our focus, are not
examined.

\paragraph{Portrait-aware retouching and region losses.} PPR10K~\cite{ppr10k2021} established
portrait retouching as a distinct task and introduced a human-region-weighted (HRP) loss that
up-weights \emph{general fidelity} inside a person mask. Our loss is related but narrower: it penalizes only the \emph{excess of one signed color direction} (green beyond
the ground truth) inside a skin mask, asymmetrically, to correct a specific diagnosed defect
rather than to raise overall regional quality. The distinction matters in practice: a symmetric regional fidelity term is tuned for overall
reconstruction quality and need not drive the diagnosed green excess to zero, whereas our
one-sided term targets exactly that excess (Sec.~\ref{sec:method}).

\paragraph{Color casts and color constancy.} Green--cyan casts are well studied in underwater
imaging, where wavelength-dependent attenuation of the \emph{scene} causes the cast and
restoration removes a physical degradation~\cite{underwater1,underwater2}. The cast we study has
a different provenance: the trained model under-restores the skin chroma that a degradation removes, a
training artifact rather than scene physics. Work on computational color constancy and white-balance
editing~\cite{afifi2019,afifi2020} likewise corrects casts present in the capture; to our
knowledge the failure mode of enhancement models \emph{leaving} a cast that surfaces on skin under a
specific training recipe has not been documented. Skin-tone-stratified evaluation follows the
precedent of fairness audits in face analysis~\cite{gendershades,skincolor2023}.

\paragraph{Synthetic degrade-and-restore training.} Synthesizing degradations over clean images
is a common route to training data at scale, from denoising and
super-resolution~\cite{realesrgan} to color enhancement. Its known risks are distribution
mismatch and over-fitting to the synthetic degradation family; we add a subtler one, a
\emph{systematic chromatic bias that surfaces on skin} in the trained model, together with a
matched FiveK/eLIR experiment isolating a training-regime effect, and a
loss-level correction compatible with the recipe's zero-annotation appeal. We evaluate on MIT-Adobe
FiveK~\cite{fivek2011}.

\section{Preliminaries and Diagnosis}\label{sec:diag}

\subsection{Subject model and training recipe}
eLIR-Net~\cite{elirnet2025} is a 6{,}795-parameter global retouching network: a \emph{BaseNet}
of three pointwise ($1{\times}1$) convolutions ($3{\to}64{\to}64{\to}3$) modulated by a
\emph{ConditionNet} that maps per-channel 16-bin RGB histograms through a shared MLP to a global
affine pair $(\alpha,\beta)$ per stage, i.e.\ pure global feature modulation,
$\tilde{x}=\alpha x + \beta$, with a one-pixel receptive field and no residual path. We use a
faithful reproduction and a channel-aware variant (\emph{v1g}, $+30$ parameters: a one-hot
channel ID concatenated to each histogram) as our working baselines; both behave identically
with respect to the defect studied here.

Following the common recipe at this scale, training is synthetically self-supervised:
155{,}792 high-quality images (DIV2K~\cite{div2k2017}, LSDIR~\cite{lsdir2023},
FFHQ~\cite{ffhq2019}) are degraded with randomized brightness
($\times0.7$--$1.3$), contrast, saturation, gamma ($0.6$--$1.4$), and per-channel white balance
($\times0.85$--$1.15$), and the network learns to restore the original under an $L_1$ + SSIM +
color-cosine + VGG objective, plus a GT-anchored luminance term (weight 1.5; Sec.~\ref{sec:bright}
sweeps this weight). No expert-retouched pairs are involved. The baseline trains with Adam
(LR $5{\times}10^{-4}\!\to\!10^{-6}$, cosine) on $256\times256$ crops; our final model resumes from it
for 150K iterations (LR $10^{-4}\!\to\!10^{-6}$, batch 32); the CSRNet-synthetic run of
Sec.~\ref{sec:exp} uses the same data and objective for 100K iterations. Full configurations
accompany the code release.

\subsection{Evaluation protocol}\label{sec:protocol}
We evaluate with a deterministic \emph{GT-anchored} protocol: a clean portrait is synthetically
degraded (brightness $\times1.45$, gamma $0.65$, saturation $0.82$; no white-balance shift), restored by the
model, and compared against the clean original. The brightness exceeds the training range and
clips the red channel on over half of skin pixels. On the 70 portraits the input skin already
shifts by $\Delta a^*$ $-9.3$, and even an exact inverse of the degradation (a reference, not a
bound) leaves $-6.9$ (baseline $-7.0$); Sec.~\ref{sec:ysweep} isolates the model's share. Metrics: whole-frame
$\mathrm{green\_shift} = \overline{\Delta G} - (\overline{\Delta R}+\overline{\Delta B})/2$ (channels in $[0,1]$) with
a flagging threshold of $0.015$ (fixed before our model was trained, then audited in
Fig.~\ref{fig:scatter}); skin-region CIEDE2000~\cite{ciede2000} and signed $\Delta a^*$ (negative $=$ green)
inside a YCbCr skin mask~\cite{skindet2002} extracted from the ground truth; PSNR; and luminance drift. The main
benchmark is 70 FFHQ portraits with skin-tone stratification (11 dark / 32 medium / 27 light),
by mean ground-truth skin luma (dark ${<}0.45$, medium $0.45$--$0.62$, light ${>}0.62$). These
portraits are drawn from FFHQ, part of the synthetic training corpus, so the protocol uses
in-distribution images (cf.\ the large-scale audit and the held-out real-photo check in
Sec.~\ref{sec:exp}); clean-input passes
serve as a control against overfitting to the degradation recipe.

\subsection{The defect: a green skin cast}
Under this protocol the synthetically-trained baselines leave a green cast in 52--54 of 70
portraits (skin $\Delta a^*$ $-7.0$ to $-8.3$), and 31--39/70 are flagged even on clean (undegraded) inputs,
where the model adds a smaller shift of its own (skin $\Delta a^*$ $-1.5$ on the 70{,}000-image
audit). On the 70-portrait degraded pass, v1g's cast is concentrated in skin (non-skin
$\Delta a^*$ $-0.80$; supplementary material). Across all baseline checkpoints in our 40-portrait audit
the mean skin $\Delta a^*$ is $-8.45$,\footnote{Per-model 70-portrait figures: $-8.28$ (eLIR-Net) and $-7.02$ (v1g).} with skin CIEDE2000 of $12.8$--$16.6$, mostly
lightness error; the $a^*$ shift alone is far above perceptibility thresholds. Whole-frame PSNR stays competitive, hiding the defect.

\subsection{Diagnosis: the cast is chrominance-intrinsic}\label{sec:ysweep}
One might attribute the defect to luminance--color entanglement, suggesting the popular
two-model decomposition: correct luminance with one network and chrominance with another in
YCbCr. For our green statistic the premise fails analytically: in a YCbCr recombination the
luminance coefficient enters $R$, $G$, and $B$ equally, so it cancels in
$\Delta G - (\Delta R{+}\Delta B)/2$, and any greenness is by construction a property of the
chrominance source alone. The cancellation is exact before clipping; in a \emph{Y-source sweep}
(four luminance candidates, including the identity, against a fixed chrominance source) through
the full pipeline the green count stays at 29/40 in every configuration, while
brightness and PSNR vary widely with the luminance choice. The cast is thus confined to the
chrominance channels, with two consequences. First, a luminance/chrominance ensemble cannot
reduce the greenness, only re-mix luminance: the two-model ensemble we evaluated measured the same
green count (29/40) as the single model it wrapped, with worse luminance drift ($0.202$ vs.\ $0.172$) and
lower PSNR ($16.76$ vs.\ $17.77$). Second, corrections partition by channel: repairing the cast acts on the chrominance channels through
the training signal (Sec.~\ref{sec:method}), whereas the brightness fix acts on luminance alone and, up to output clipping, leaves the repaired $C_bC_r$ unchanged (Sec.~\ref{sec:bright}).

\paragraph{Why green, and why skin.} An eval-only decomposition of the 70{,}000-image audit
(supplementary material) shows that the model under-restores skin chroma lost to the degradation
and leaves skin hue rotated toward yellow in every condition, clean inputs included; both lower
skin $a^*$. Without clipping ($\times1.00$) its output nearly keeps the input's skin $\Delta a^*$ ($-6.2$
vs.\ $-6.5$). On unclipped pixels at $\times1.45$ the degradation lowers mean skin chroma (ratio $0.82$)
but not that of non-skin chromatic pixels ($1.04$), while the model's gain is similar on both
($1.14$ vs.\ $1.13$). Per-tone effects differ between $\times1.45$ and $\times1.00$ (exploratory).
Sec.~\ref{sec:exp} ties the \emph{origin} to the recipe.

\section{Method}\label{sec:method}

\subsection{Analysis of failing objectives}
We first explain why several natural objectives leave the skin cast intact.
\emph{Input-anchored} color/luminance preservation losses (penalizing deviation of the
output's green ratio or luma from the \emph{input}) directly oppose the correction the
network must perform, and never reference the ground truth against which the cast is
measured. More subtly, a \emph{GT-anchored global-mean} green loss (penalizing the
image-level mean green residual $\bar g=\mathrm{mean}(\hat y-y)_G-\tfrac12[\mathrm{mean}(\hat y-y)_R+\mathrm{mean}(\hat y-y)_B]$)
also fails: a globally-modulated $1{\times}1$ network satisfies an image mean by
\emph{regional offsetting} (adding magenta in the background to cancel green on skin),
so the mean vanishes while skin stays green. Empirically, this objective worsens
the defect (portrait green-tint rate $52\to57$ of $70$; Sec.~\ref{sec:exp}).
Architecture-side variants within this family did not remove the cast either: enriching or
replacing the conditioning signal (22-bin, YCbCr, and dual RGB+YCbCr histograms) and adding
residual modulation, the v1g-22, residual, YCbCr-hist, and dual-hist variants of Fig.~\ref{fig:scatter}, still produced
green casts in 29--34 of 40 audited portraits, consistent with an origin in the
training signal rather than the conditioning detail. The complete set of attempted
alternatives, including the luminance/chrominance ensembles of Sec.~\ref{sec:ysweep} and sweeps
over input-anchored color-preservation weights, is reported in the supplementary material.

\subsection{Skin-masked asymmetric green-excess loss}
\begin{figure}[tb]\centering
\resizebox{\linewidth}{!}{%
\begin{tikzpicture}[font=\footnotesize,>=Stealth,
  box/.style={draw,rounded corners,align=center,minimum height=8mm,inner sep=3pt,fill=blue!4},
  mbox/.style={draw,rounded corners,align=center,minimum height=8mm,inner sep=3pt,fill=orange!10},
  op/.style={draw,circle,inner sep=1.5pt,fill=gray!12}]
\node[box] (pred) {$\hat y$};
\node[box,below=5mm of pred] (gt) {$y$};
\node[box,right=7mm of $(pred)!0.5!(gt)$] (res) {residual\\$r=\hat y - y$};
\node[box,right=8mm of res] (ge) {green excess\\$g=r_G-\tfrac12(r_R{+}r_B)$};
\node[box,right=8mm of ge] (relu) {asymmetric\\ReLU $\max(g,0)$};
\node[op,right=8mm of relu] (mult) {$\odot$};
\node[box,right=7mm of mult] (loss) {masked mean\\$\mathcal{L}_{\text{skin}}$};
\node[mbox,below=8mm of relu] (mask) {YCbCr skin mask $M$\\(derived from $y$)};
\draw[->](pred)--(res); \draw[->](gt)--(res);
\draw[->](res)--(ge); \draw[->](ge)--(relu); \draw[->](relu)--(mult); \draw[->](mult)--(loss);
\draw[->](mask)-|(mult);
\draw[->](gt.south)|-(mask.west);
\node[above=2mm of relu,font=\scriptsize,text=gray] {penalizes only $g{>}0$};
\end{tikzpicture}}
\caption{Skin-masked asymmetric green-excess loss. The per-pixel green residual $g$ of
$\hat y - y$ is passed through a ReLU (penalized only for $g{>}0$), weighted by a
GT-derived skin mask $M$, and averaged. Mask-localized gradients prevent the regional
offsetting of global-mean losses, and the ReLU penalizes only excess green.}
\label{fig:method}
\end{figure}
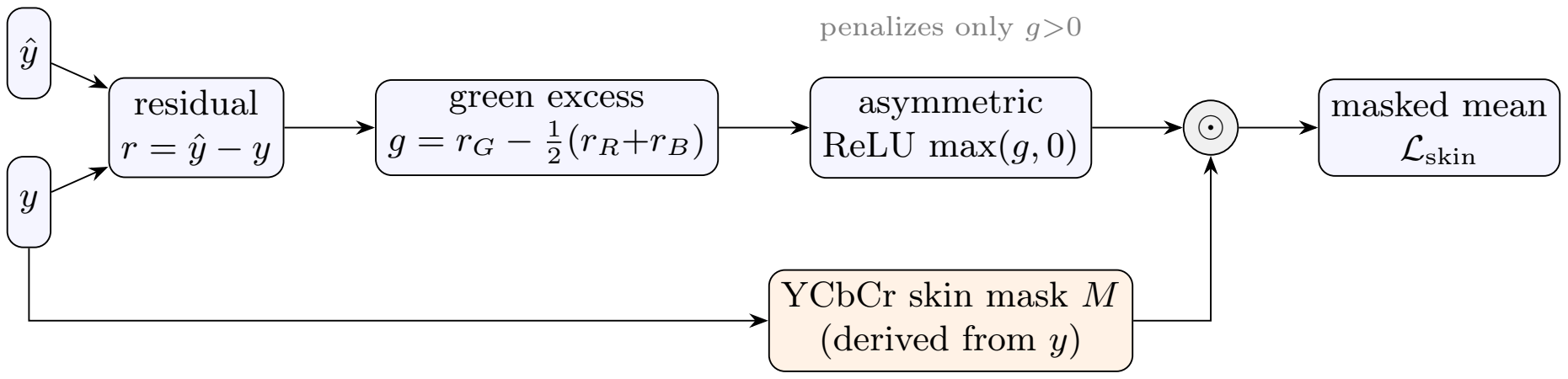

To address this spatially localized failure, we derive a binary skin mask
$M$ from the \emph{ground-truth} image via a fixed YCbCr rule
($77\!\le\!C_b\!\le\!127,\ 133\!\le\!C_r\!\le\!173$ on $[0,255]$, luma $>0.2$), and penalize,
at every skin pixel, only the \emph{excess} green of the per-pixel residual relative to GT:
\begin{equation}
g_{p} = (\hat y-y)_G^{p} - \tfrac12\big[(\hat y-y)_R^{p}+(\hat y-y)_B^{p}\big],\qquad
\mathcal{L}_{\text{skin}} = \frac{\sum_p M_p\,\mathrm{ReLU}(g_p)}{\sum_p M_p + \epsilon}.
\end{equation}
The loss involves three design choices. (i) \textbf{Per-pixel, mask-weighted}: dense gradients on
skin pixels prevent the regional offsetting that satisfies global means.
(ii) \textbf{Asymmetric} ($\mathrm{ReLU}$): only green \emph{above} GT is penalized, so the
term's direct gradient is one-sided and never rewards adding green. (In combination with the
base reconstruction objective it can still overshoot toward magenta when the base model is
already near-neutral, e.g.\ under real-pair training; Sec.~\ref{sec:lim}.) (iii) \textbf{GT-anchored}: the target
is the correct color, aligning the term with reconstruction rather than opposing it.
The loss adds no parameters and is applied on top of the baseline reconstruction objective
with a single weight $\tau$; we use $\tau{=}5$ (Sec.~\ref{sec:exp} ablates $\tau$). The mask and
ground truth enter only during training; the deployed model is the unchanged global $1{\times}1$
operator and requires no skin mask, face detector, or ground truth at inference.
Fig.~\ref{fig:method} summarizes the computation.

\section{Experiments}\label{sec:exp}

\paragraph{Protocol.} Unless noted, we use the 70-portrait protocol of Sec.~\ref{sec:protocol} and report
the green-tint rate (whole-frame green\_shift $>0.015$), skin $\Delta a^*$ (${<}0$ is green) and skin
CIEDE2000 on the GT-derived skin mask.

\begin{figure}[tb]\centering
\includegraphics[width=\linewidth]{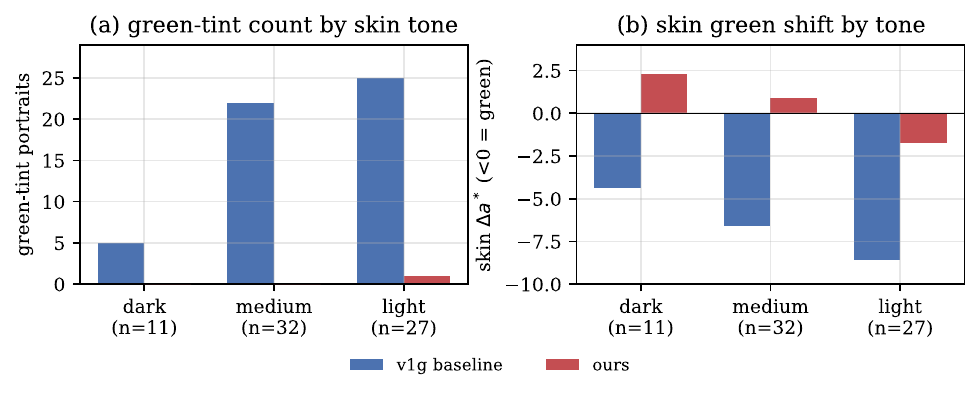}
\caption{Skin-tone stratification: green-tinted portraits per stratum (a; counts) and mean skin
$\Delta a^*$ (b) for the v1g baseline vs.\ ours; the cast is suppressed in
every stratum.}
\label{fig:strat}
\end{figure}

\paragraph{Main result.} Table~\ref{tab:main} and Fig.~\ref{fig:strat} show our loss suppresses
the cast: the green-tint rate drops from $52/70$ to $1/70$ and mean skin $\Delta a^*$ moves
from $-7.02$ (clearly green) to $+0.13$ (near zero on average), while skin CIEDE2000 barely
changes ($13.17\!\to\!12.91$): the loss cancels the mean green--magenta bias (unclipped skin turns magenta), but the lightness
error persists and the $b^*$ error grows (supplementary material). The correction holds across stratified skin tones (dark $5\!\to\!0$,
medium $22\!\to\!0$, light $25\!\to\!1$; Fig.~\ref{fig:strat}).
Fig.~\ref{fig:teaser} and the supplementary material show representative corrections across skin tones and cast severities.

\paragraph{Large-scale audit.} To confirm the effect is not an artifact of the 70-image
sample, we repeat the identical measurement on all 70{,}000 FFHQ images. Because FFHQ is part
of the synthetic training corpus, we report this as a large-scale \emph{distribution audit}
rather than a held-out test, and make no generalization claim beyond the training distribution.
The operator is nonetheless a global color map that receives no explicit identity or
spatial-content signal beyond a spatially pooled color histogram; the audit
therefore probes the systematic behavior of the learned global color policy, and the
clean-input control below (undegraded inputs) further guards against overfitting to the
synthetic degradation.
The trend holds at scale: the baseline flags a green cast on $47{,}700/70{,}000$ portraits
($68.1\%$, Wilson $95\%$ CI $[67.8,68.5]\%$; mean skin $\Delta a^*=-6.50$), which our loss
reduces to $1{,}510/70{,}000$ ($2.2\%$, CI $[2.1,2.3]\%$; $\Delta a^*=+0.80$) at identical
parameter count. On already-clean inputs (a clean-input control) the baseline still flags green on
$42.7\%$ of portraits, versus $2.1\%$ for ours; here the correction trades the baseline's mild
green (skin $\Delta a^*$ $-1.5$) for a magenta shift of larger magnitude ($+4.4$) at
higher skin CIEDE2000 ($5.6\!\to\!6.4$): a directional exchange, so signed skin
$\Delta a^*$ must accompany green-flag counts. Per-stratum results at the 70{,}000 scale
(baseline cast significant in every stratum; ours ${\le}4.0\%$) are given in the supplementary
material.

\paragraph{Held-out real-photo check.} To test whether the effect is tied to the training
corpus, we repeat the protocol on 500 CelebA-HQ portraits~\cite{celebamaskhq2020}, real
photographs verified disjoint from all 155{,}792 training images by perceptual hashing. The
cast and correction transfer with comparable magnitude: the baseline flags $64.4\%$ (Wilson $95\%$ CI $[60.1,68.5]\%$; mean skin $\Delta a^*$
$-6.95$) and ours $0.8\%$ ($[0.3,2.0]\%$; $+0.63$), closely matching the in-distribution audit,
and the clean-input exchange also transfers ($-2.08 \to +3.82$ at skin CIEDE2000
$5.20\!\to\!6.38$). Re-scoring the same outputs with the dataset's hand-annotated skin parsing
masks in place of our YCbCr rule leaves the conclusions unchanged (baseline $-6.79$, ours
$+0.54$; Spearman $0.95/0.96$ against the YCbCr-mask scores, mask IoU $0.85$), so the
measurement is not an artifact of the mask rule. A second real-photo source agrees: on
zero-shot FiveK the baseline renders skin green ($\Delta a^*$ $-4.48$ over the 490
skin-detected test images) while ours is near-neutral ($-0.96$).

\begin{table}[t]
\centering
\caption{Main result on 70 FFHQ portraits (GT-anchored). The skin-masked loss suppresses the
green cast at zero parameter cost; skin CIEDE2000 barely changes and whole-frame PSNR on this
synthetic stress test decreases slightly ($17.39\!\to\!16.90$); standard-benchmark PSNR/SSIM
is preserved (Table~\ref{tab:fivek}).}
\label{tab:main}
\begin{tabular}{lccccc}
\toprule
Model & params & green-tint$\downarrow$ & skin $\Delta a^*$ & skin CIEDE2000$\downarrow$ & PSNR \\
\midrule
eLIR-Net (v1g) baseline & 6{,}825 & 52/70 & $-7.02$ & 13.17 & 17.39 \\
\ +\,global-mean green loss & 6{,}825 & 57/70 & $-7.73$ & 13.26 & 17.61 \\
\textbf{\ +\,skin green loss (ours)} & 6{,}825 & \textbf{1/70} & \textbf{$+0.13$} & \textbf{12.91} & 16.90 \\
\bottomrule
\end{tabular}
\end{table}

\paragraph{Loss weight and loss form.} $\tau{=}1/2/5$ yields $17/8/1$ flags (degraded skin
$\Delta a^*$ $-3.40/{-}1.56/{+}0.13$) while clean-input over-correction grows ($+1.4/{+}2.7/{+}4.2$);
$\tau{=}10$ over-corrects (degraded skin $\Delta a^*$ $+2.10$). We use $\tau{=}5$. At matched, gradient-calibrated
coefficients, symmetric green and Lab-$a^*$ objectives and PPR10K-form skin-weighted MSE leave
the cast largely intact (39--59 of 70; supplementary material).

\begin{figure}[tb]\centering
\includegraphics[width=.6\linewidth]{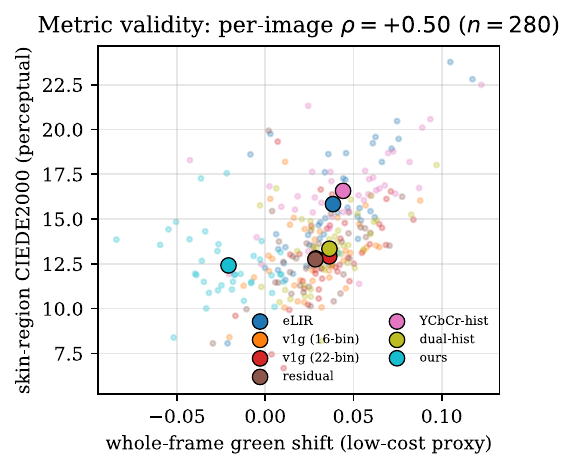}
\caption{The low-cost whole-frame green metric tracks perceptual skin CIEDE2000 (per-image
$\rho=0.50$, $n=280$). Bold markers are per-method means; faint points are per-image.
Methods (all synthetically trained): the eLIR reproduction; the channel-aware baseline with
16- and 22-bin histograms (v1g-16/-22, Sec.~\ref{sec:diag}); three diagnostic variants
(residual: residual modulation; YCbCr-hist: YCbCr-histogram conditioning; dual-hist: dual
RGB+YCbCr histograms); and ours (the v1g-16 architecture retrained with our loss, Sec.~\ref{sec:method}).}
\label{fig:scatter}
\end{figure}

\paragraph{Metric audit.} We validate the low-cost whole-frame green\_shift against
perceptual skin CIEDE2000 (Fig.~\ref{fig:scatter}). At the method level the two metrics rank consistently (Spearman $\rho{=}0.89$); the per-image correlation is
weaker but positive ($\rho{=}{+}0.50$, $n{=}280$; the seven synthetically-trained checkpoints
of Fig.~\ref{fig:scatter} over the 40-portrait audit), reflecting
per-image chroma variance not captured by a green-only statistic. However, green\_shift does not
flag magenta over-correction, so we report it paired with skin $\Delta a^*$. The separation
holds for every flag threshold in $[0.005,0.030]$ on all evaluation sets (supplementary material).

\paragraph{Cross-architecture check.}
To ask whether the cast is a single-model artifact, we run CSRNet~\cite{csrnet2020} (36.5K
params, a CNN-conditioned global retouching net trained on real expert-retouched pairs)
through the same portrait protocol. CSRNet does \emph{not} reproduce the green cast: it renders
skin neutrally in its own FiveK domain (skin $\Delta a^*\!=\!+0.03$ over 490 images) and shows
no green on our portraits; its out-of-distribution behavior there trends magenta, a
different and opposite bias. This is consistent with a
training origin rather than an architectural one; but CSRNet also differs in \emph{training
data}, so it alone cannot isolate the cause. The next experiment addresses this directly.

\paragraph{A $2\times2$ recipe-localization test.}
To localize the source of the cast we cross \emph{two architectures} (eLIR, histogram-conditioned; and
CSRNet~\cite{csrnet2020}, CNN-conditioned) with \emph{two training regimes} (the synthetic
self-supervised protocol; and real paired data, MIT-Adobe FiveK), and measure skin
$\Delta a^*$, each cell within its own image domain (Table~\ref{tab:causal}). In this
observational $2\times2$ the cast tracks the recipe: both architectures leave skin green under
synthetic self-supervision (skin $\Delta a^*$ $-8.3$ / $-2.9$) and both are near-neutral, each
in its own image domain, on real pairs ($-0.96$ / $+0.03$); on the same degraded portraits the
synthetically trained residuals are also the greener ones ($-8.28$ vs.\ $-4.44$; $-2.93$ vs.\ $+2.37$).
Because the cells also differ in corpus, the controlled evidence is a same-corpus control (FiveK
corpus, targets, architecture, schedule, and seeds fixed; only the input-generation regime
varied): the synthetic cell is greener by ${\approx}3.7$ $\Delta a^*$ on identical inputs in all three
seeds, though a single-seed PPR10K-derived control reversed the sign (supplementary material).
Together, these results show that the cast is not specific to one architecture and, in the matched
FiveK/eLIR setting, tracks the \emph{synthetic self-supervised degradation--correction protocol}. Recipe localization is a statement about
\emph{in-domain} behavior, the regime in which a deployed model operates. Out of their
training domain, both real-pair-trained models drift in architecture-specific directions on
the degraded portraits (eLIR stays green, CSRNet overshoots to magenta, from an input at $-9.3$), so the
FiveK cell's neutrality is an in-domain statement, not a claim
of out-of-distribution robustness. Real-pair training is therefore not a practical substitute:
paired data seldom covers the degradations a deployed model must undo; our loss instead suppresses the cast
while keeping the recipe's zero-annotation appeal. The severity difference across architectures is real but
unattributed: the two models differ simultaneously in conditioning, capacity, and
structure, so our design cannot isolate its source. Training in-domain on FiveK
also resolves the absolute-dB scale (eLIR reaches
$22.4$\,dB), and adding our loss during FiveK training adds no PSNR/SSIM cost (PSNR $+0.13$, SSIM $+0.004$); the model's chromatic error rises slightly (whole-frame CIEDE2000 $8.77\!\to\!9.69$), consistent with the loss targeting the green-cast regime rather than near-neutral models. The
corrective loss is not specific to the eLIR family: trained on the synthetic recipe, both a
CNN-conditioned network (CSRNet) and a sequential-operator network (NeurOp) develop the cast
(portrait green-tint $23$ and $52$ of $70$), and our loss suppresses it in each (to ${\le}6\%$ on
portraits and ${\le}2.2\%$ on held-out CelebA-HQ; skin $\Delta a^*$ to near-zero), with the same
clean-input magenta trade-off (Table~\ref{tab:archmain}; full per-architecture metrics in
the supplementary material).

\begin{table}[t]
\centering
\caption{$2\times2$ recipe-localization test (observational; the same-corpus control of
Sec.~\ref{sec:exp} provides the controlled evidence): skin $\Delta a^*$ (${<}0$ is green), each cell
probed within its own image domain. Two different architectures both leave skin green under the
synthetic self-supervised protocol and both are near-neutral trained on real pairs.}
\label{tab:causal}
\begin{tabular*}{\linewidth}{@{\extracolsep{\fill}}lcc@{}}
\toprule
Architecture & synthetic protocol & real FiveK pairs \\
\midrule
eLIR-Net (histogram-cond., 6.8K) & $-8.28$ (green) & $-0.96$ (neutral) \\
CSRNet (CNN-cond., 36.5K) & $-2.93$ (green) & $+0.03$ (neutral) \\
\bottomrule
\end{tabular*}
\par\smallskip
{\footnotesize\raggedright Domains: synthetic column = 70-portrait GT-anchored protocol (exact-inverse
reference $-6.9$, Sec.~\ref{sec:protocol}); real column =
FiveK test (the 490 of 500 test images with detectable skin). eLIR-FiveK is our 40K-iter run;
CSRNet-FiveK is the official checkpoint; CSRNet-synthetic is our run (100K iters,
Sec.~\ref{sec:diag}).\par}
\end{table}

\begin{table}[t]
\centering
\caption{The same loss suppresses the cast across three ultra-light architectures with distinct
conditioning mechanisms (synthetic-trained $\to$ +loss; skin $\Delta a^*$ to near-neutral in all
cases; per-architecture $\Delta a^*$ in the supplementary material). 70-proto: green count of 70;
CelebA-HQ: held-out real-photo rate.}
\label{tab:archmain}
\setlength{\tabcolsep}{5pt}
{\small
\begin{tabular}{llrcc}
\toprule
architecture & mechanism & params & 70-proto (green) & CelebA-HQ (rate) \\
\midrule
eLIR (v1g) & histogram-cond. & 6{,}825 & $52/70 \to 1/70$ & $64.4\% \to 0.8\%$ \\
CSRNet & CNN-cond. & 36{,}489 & $23/70 \to 2/70$ & $21.4\% \to 0.4\%$ \\
NeurOp & sequential op. & 28{,}108 & $52/70 \to 4/70$ & $52.0\% \to 2.2\%$ \\
\bottomrule
\end{tabular}}
\end{table}

\paragraph{Fidelity on a standard benchmark.}
We check that suppressing the cast does not harm standard-benchmark fidelity. Evaluated
\emph{zero-shot} on MIT-Adobe FiveK (models trained on synthetic data, not FiveK),
Table~\ref{tab:fivek} shows our loss preserves fidelity (PSNR $+0.27$ vs.\ the eLIR
reproduction, $+0.77$ vs.\ v1g; SSIM and CIEDE2000 also improve), and the inference luma gain (Sec.~\ref{sec:bright}) adds
$+2.3$\,dB where inputs are under-exposed. Absolute values are modest because the comparison
is zero-shot cross-dataset; when the same architecture is trained \emph{in-domain} on FiveK it
reaches $22.4$\,dB (Table~\ref{tab:fivek}); the original eLIR-Net reports $24.85$\,dB with its own
FiveK training setup, consistent with the reported range for this
family of models trained on paired data.

\begin{table}[!htb]
\centering
\caption{Fidelity on MIT-Adobe FiveK (expert C, 500-image test; whole-frame CIEDE2000). Top:
\emph{zero-shot} models trained on synthetic degradation; the skin-masked loss preserves fidelity
and the luma gain adds $+2.3$\,dB on under-exposed inputs (a fixed gain harms clean inputs;
Sec.~\ref{sec:bright}). Bottom: \emph{in-domain} FiveK-trained (40K iters), resolving the
absolute-dB scale.}
\label{tab:fivek}
\begin{tabular}{lcccc}
\toprule
Model & PSNR & SSIM & CIEDE2000 & green\_shift \\
\midrule
eLIR baseline (reproduction) & 13.65 & 0.6408 & 20.52 & $-0.0210$ \\
v1g (channel-aware) & 13.15 & 0.6268 & 20.99 & $-0.0116$ \\
Ours (final) & \textbf{13.92} & \textbf{0.6573} & \textbf{20.48} & $-0.0282$ \\
Ours $+$ luma gain $1.8$ & \textbf{16.22} & \textbf{0.7182} & \textbf{17.06} & $-0.0277$ \\
\midrule
\multicolumn{5}{l}{\emph{FiveK-trained (in-domain, 40K iters)}} \\
eLIR-Net ($+$orig.\ loss) & 22.44 & 0.8673 & 8.77 & --- \\
eLIR-Net ($+$ours) & 22.57 & 0.8716 & 9.69 & --- \\
CSRNet~\cite{csrnet2020} (ref.) & 23.58 & 0.8820 & 7.84 & --- \\
\bottomrule
\end{tabular}
\end{table}

\section{Brightness: Capacity Limit and Inference-Time Control}\label{sec:bright}
A second failure under this protocol is exposure: the model recovers only
${\sim}55\%$ of the required luminance correction. Unlike the green cast, simple loss
reweighting did not resolve it in our pointwise global-affine architecture. Raising the
GT-anchored luminance loss weight from 1.5 to 5.0 leaves the mean
luminance error essentially unmoved ($0.122 \to 0.115$), whereas an
\emph{oracle} that projects the output onto the ground-truth luma recovers $+9$\,dB. The gap is
consistent with a capacity limit rather than a deficient objective: a pointwise global-affine
model cannot recover clipped highlights and has no spatial context, so the achievable luminance
correction saturates in our sweeps, the same root cause as the architecture's known
over-exposure failure.

We recover this gap at inference time by amplifying the luminance correction the model
already makes:
\begin{equation}
\mathrm{out} \mathrel{+}= (g-1)\,(Y_{\mathrm{out}} - Y_{\mathrm{in}}),
\end{equation}
added equally to all channels so that only luma moves; up to output clipping, $C_bC_r$ (hence
green\_shift) is unchanged. At $g{=}1.8$ on the over-exposed protocol inputs, mean
luminance error drops $0.117 \to 0.033$ and PSNR rises $+5.0$\,dB ($16.90 \to 21.90$) with the
green count unchanged (1/70); $g{=}2.5$ over-amplifies (18.93\,dB). On zero-shot FiveK, whose
inputs are systematically under-exposed, $g{=}1.8$ adds $+2.30$\,dB (Sec.~\ref{sec:exp}).

This control is not automatic. A fixed gain degrades
clean inputs ($25.27 \to 21.06$\,dB), and an input-brightness-adaptive gate cannot distinguish
``bright but correct'' from ``over-exposed'' using the global statistics available to this model
class ($25.3 \to 23.3$ on clean). We therefore expose it as a manual inference-time control
(default $g{=}1.0$, recommended $1.5$--$1.8$ for exposure-degraded input); when to apply it is
left to the caller.

\section{Limitations}\label{sec:lim}
\textbf{Synthetic evaluation degradations.} Our portrait protocol degrades with
brightness/gamma/saturation only, with brightness beyond the training range, while training used
white-balance jitter; absolute restoration numbers do not transfer to arbitrary real degradations.
\textbf{Scope of recipe localization.} The cross-architecture evidence is observational; the
controlled regime effect holds for eLIR on FiveK, and a single-seed PPR10K-derived control reversed it. Recipe
localization is an in-domain statement; out-of-distribution color robustness
is a separate open problem, as real-pair models drift out of domain (Sec.~\ref{sec:exp}). Whether the recipe
induces the same bias across the broader space of enhancement architectures and
synthetic-degradation families remains open, as does why the recipe yields a model that under-restores skin chroma (Sec.~\ref{sec:ysweep}); severity also varies with architecture for reasons our two-model design cannot isolate, and a
1.77M-parameter DeepLPF trained with the same recipe retained the cast at reduced magnitude
(supplementary material). \textbf{Loss regime.} The asymmetric green-excess loss targets the signed
green--magenta axis, not total skin-color fidelity, acting as a near-constant shift of skin $a^*$
(supplementary material), and is designed for the green-cast regime;
in the near-neutral real-pair regime it over-corrects skin toward magenta and total skin color
error worsens (skin $\Delta a^*$ $+4.48$, skin-region CIEDE2000 $9.68 \to 11.07$), the direction our
audited low-cost metric is blind to (Sec.~\ref{sec:exp}), so we pair it with signed $\Delta a^*$ and
recommend the loss specifically for synthetic-protocol training; its effectiveness is
demonstrated on three architectures (eLIR, CSRNet, NeurOp), and whether it extends beyond these
remains open. \textbf{Data skew and strata.} FFHQ under-represents dark skin (11/70); flags after correction are
low in every stratum (0/11, 0/32, 1/27), but on clean inputs dark skin ends most magenta
($+0.9\!\to\!+6.0$; supplementary material), per-stratum samples are small, and the YCbCr rule covers the darkest
tones least well (only partly controlled by the CelebA-HQ parsing-mask agreement, Sec.~\ref{sec:exp}).
Strata are defined by observed skin luma, a lighting-confounded proxy rather than a skin-tone
scale such as ITA~\cite{ita1991}, Fitzpatrick~\cite{fitzpatrick1988}, or Monk~\cite{monk2023};
scale-based stratification is future work. \textbf{Manual exposure control.} The luma-delta gain requires the application to know the
input is exposure-degraded; automating this is beyond the model class (Sec.~\ref{sec:bright}). \textbf{Data and intent.} DIV2K, LSDIR, FFHQ, CelebA-HQ~\cite{celebamaskhq2020}, MIT-Adobe
FiveK~\cite{fivek2011} and PPR10K~\cite{ppr10k2021} are public datasets used under their research terms; no new
human data were collected. Skin masks derive from
ground-truth chroma thresholds or the datasets' published annotations, not identity attributes;
the loss penalizes only green in excess of each subject's own ground-truth skin tone,
and is not designed to standardize, lighten, or beautify skin.

\section{Conclusion}\label{sec:conc}
We documented a systematic failure mode of the common synthetic self-supervised recipe for
ultra-lightweight photo retouching: the trained model leaves a green chrominance cast on human
skin that whole-frame PSNR/SSIM do not capture. The cast is chrominance-intrinsic, recurs across
two architectures and, in a same-corpus
FiveK control, tracks the training recipe; a
per-pixel, skin-masked, asymmetric green-excess loss suppresses it (52/70 $\to$ 1/70 across skin
tones at zero parameter cost) with in-domain FiveK PSNR/SSIM unchanged and a documented
clean-input magenta trade-off. We audited a low-cost whole-frame green metric (method-level rank
correlation $0.89$ with perceptual skin color difference, a moderate per-image correlation, and a
magenta blind spot), and converted the remaining brightness deficit into a chrominance-preserving
inference-time control ($+5.0$\,dB on exposure-degraded inputs;
a fixed gain harms clean inputs, $25.27\!\to\!21.06$\,dB). The result is a single
6{,}825-parameter model in 35\,KB of TFLite. More broadly, models trained on synthesized
degradations should have skin-region color evaluated with its own metric, since whole-frame
fidelity scores do not expose such casts.

\bibliographystyle{splncs04}
\bibliography{refs}

\clearpage
\setcounter{section}{0}\setcounter{table}{0}\setcounter{figure}{0}\setcounter{footnote}{0}
\renewcommand{\thesection}{S\arabic{section}}
\renewcommand{\thetable}{S\arabic{table}}
\renewcommand{\thefigure}{S\arabic{figure}}
\renewcommand{\theHsection}{S\arabic{section}}
\renewcommand{\theHtable}{S\arabic{table}}
\renewcommand{\theHfigure}{S\arabic{figure}}
\begin{center}{\Large\bfseries Supplementary Material\par}\end{center}
\vspace{1ex}
\noindent This supplement provides the full evaluations summarized in the main paper:
held-out real-photo results (Sec.~\ref{S:sec:celeba}, backing main-paper Sec.~5), the
full-corpus stratified audit (Sec.~\ref{S:sec:strata}), the three-architecture generalization
of the loss (Sec.~\ref{S:sec:arch}, backing main-paper Sec.~5), the failed-remedy and
input-anchored sweeps (Sec.~\ref{S:sec:failed}, backing main-paper Sec.~4.1), an
out-of-distribution synthetic probe (Sec.~\ref{S:sec:ood}), statistical-robustness tests
(Sec.~\ref{S:sec:stats}), the same-corpus training-regime control (Sec.~\ref{S:sec:regime}),
regional and symmetric loss ablations (Sec.~\ref{S:sec:lossabl}), threshold sensitivity of the
green flag (Sec.~\ref{S:sec:threshold}), skin vs.\ non-skin, degradation-jitter and capacity controls
(Sec.~\ref{S:sec:controls}), an eval-only decomposition of why the shift is green and appears on
skin, added at camera-ready (Sec.~\ref{S:sec:mechanism}), on-device latency (Sec.~\ref{S:sec:latency}), additional qualitative
examples (Sec.~\ref{S:sec:qual}), the planned code release (Sec.~\ref{S:sec:repro}), and image credits (Sec.~\ref{S:sec:credits}). Section and table
references without ``main paper'' point
within this document. The eLIR (v1g) baseline and final checkpoints, and all thresholds, are
frozen. Every evaluation here uses them unchanged, except the model comparisons of
Sec.~\ref{S:sec:arch} (loss generalization), Sec.~\ref{S:sec:failed} (failed remedies), and
Secs.~\ref{S:sec:regime}, \ref{S:sec:lossabl} and \ref{S:sec:controls} (controls added after review), which
train the alternative models they report.

\section{Held-Out Real-Photo Evaluation (CelebA-HQ)}
\label{S:sec:celeba}
Main-paper Sec.~5 presents a summary of this evaluation; the full protocol and tables follow.

\paragraph{Disjointness protocol.}
We evaluate on 500 real CelebA-HQ portraits (1024$^2$, aligned as in FFHQ), sampled with a
fixed seed. To justify the held-out designation we verified that the 500 evaluation images are
disjoint from the 155{,}792 training images by perceptual hashing (dHash): there are
\emph{zero} exact hash collisions, and of the 24 pairs at Hamming distance 2--3, 12
were inspected manually and all show different people (different source identities; no shared
photograph). The source (CelebA) is distinct from the Flickr-based training corpus, so this is
a cross-source generalization test. A 70-portrait sanity re-run on this pipeline reproduces the
main-paper diagnostic exactly (52/70, mean skin $\Delta a^*$ $-7.02$).

\paragraph{Results.}
Table~\ref{S:tab:celeba} gives the full evaluation. The degraded-input cast transfers at nearly
the same strength as on the training distribution and is suppressed to near-neutral by our loss.
The clean-input ``directional exchange'' described in main-paper Sec.~5 also transfers: the
residual green of the baseline ($-2.08$) becomes a larger magenta shift for our model ($+3.82$).
Table~\ref{S:tab:corr} places the held-out numbers beside the in-distribution FFHQ corpus.

\begin{table}[t]
\centering
\caption{Held-out CelebA-HQ real photos ($n=500$). Rates with Wilson 95\% intervals; mean skin
$\Delta a^*$ with bootstrap 95\% intervals. Negative $\Delta a^*$ is green, positive is magenta.}
\label{S:tab:celeba}
\setlength{\tabcolsep}{6pt}
\begin{adjustbox}{max width=\textwidth,center}%
\begin{tabular}{llcccc}
\toprule
pass & model & green rate [Wilson95] & skin $\Delta a^*$ [boot95] & skin CIEDE2000 & PSNR \\
\midrule
degraded & baseline & 64.4\% [60.1, 68.5] & $-6.95$ [$-7.23$, $-6.67$] & 13.14 & 17.28 \\
degraded & \textbf{ours} & \textbf{0.8\%} [0.3, 2.0] & $\mathbf{+0.63}$ [$+0.37$, $+0.91$] & 13.38 & 16.87 \\
\midrule
clean & baseline & 43.2\% [38.9, 47.6] & $-2.08$ [$-2.34$, $-1.81$] & 5.20 & 26.94 \\
clean & \textbf{ours} & \textbf{1.0\%} [0.4, 2.3] & $+3.82$ [$+3.55$, $+4.11$] & 6.38 & 26.84 \\
\bottomrule
\end{tabular}%
\end{adjustbox}
\end{table}

\begin{table}[t]
\centering
\caption{Held-out CelebA-HQ vs.\ the in-distribution 70{,}000-image FFHQ corpus (green rate,
mean skin $\Delta a^*$). The cast transfers at essentially the same strength.}
\label{S:tab:corr}
\setlength{\tabcolsep}{8pt}
\begin{adjustbox}{max width=\textwidth,center}%
\begin{tabular}{lcc}
\toprule
 & FFHQ 70K (in-distribution) & CelebA-HQ 500 (held-out) \\
\midrule
baseline, degraded & 68.1\%, $-6.50$ & 64.4\%, $-6.95$ \\
ours, degraded     & 2.2\%, $+0.80$  & 0.8\%, $+0.63$ \\
baseline, clean    & 42.7\%, $-1.46$ & 43.2\%, $-2.08$ \\
ours, clean        & 2.1\%, $+4.37$  & 1.0\%, $+3.82$ \\
\bottomrule
\end{tabular}%
\end{adjustbox}
\end{table}

\paragraph{Mask cross-validation.}
Because our loss and our metric both use a YCbCr skin box, we test whether the result
is an artifact of that mask. We re-scored all 500 images with the dataset's manually annotated
``skin'' parsing masks (19-class hand labels, independent of our box), changing \emph{only} the
scoring mask (model, loss, and protocol unchanged). The conclusion is unchanged
(Table~\ref{S:tab:maskcv}): per-image $\Delta a^*$ under the two masks correlates at Pearson
$0.93$/$0.95$ and Spearman $0.95$/$0.96$, with mask IoU $0.847$. The finding is therefore largely
not an artifact of the mask rule, though the YCbCr box covers the darkest tones least well and
this aggregate cross-validation does not stratify by tone (main-paper Sec.~7).

\begin{table}[t]
\centering
\caption{Mask cross-validation on CelebA-HQ ($n=500$): our YCbCr box vs.\ manual parsing masks.
The absolute $\Delta a^*$ here is from this separate re-scoring run; ours reads $+0.61$ vs.\
Table~\ref{S:tab:celeba}'s $+0.63$.}
\label{S:tab:maskcv}
\setlength{\tabcolsep}{8pt}
\begin{adjustbox}{max width=\textwidth,center}%
\begin{tabular}{lccc}
\toprule
 & YCbCr box & manual parsing & correlation (Pearson / Spearman) \\
\midrule
baseline skin $\Delta a^*$ & $-6.95$ & $-6.79$ & $0.93$ / $0.95$ \\
ours skin $\Delta a^*$     & $+0.61$ & $+0.54$ & $0.95$ / $0.96$ \\
\bottomrule
\end{tabular}%
\end{adjustbox}
\end{table}

\paragraph{Residual cases and licensing.}
Four of the 500 corrected images retain a residual green flag; their identifiers and metrics will
be included in the code release, but we do not reproduce CelebA-HQ face panels here, as the
dataset's license restricts redistribution. All qualitative face panels in this supplement use
FFHQ images only, matching the main paper, and are used for non-commercial research. The FFHQ
dataset is distributed under CC BY-NC-SA 4.0, while each FFHQ image keeps its original Flickr
license; Sec.~\ref{S:sec:credits} credits the author, source and license of each displayed image, which
we cropped and color-processed. Displayed faces are FFHQ subjects that passed our age
screening: images 20719, 36016, 57991 (main-paper Fig.~1), 40403, 16976, 21782, 11340, 17475, 61228
(Fig.~\ref{S:fig:qual}), and 62895, 62430, 28568 (Fig.~\ref{S:fig:residual}). All data are public
research datasets: DIV2K and LSDIR (training only), FFHQ, CelebA-HQ with the CelebAMask-HQ parsing
masks, MIT-Adobe FiveK, a PPR10K-derived subset (Sec.~\ref{S:sec:regime}), and the synthetic PFHQ faces
(Sec.~\ref{S:sec:ood}). These public datasets include images of minors; such images enter only
aggregate statistics and are never displayed. No new human data were collected, and we did not
obtain separate consent from the depicted individuals for this study.

\section{Full-Corpus Stratified Audit (70{,}000 FFHQ portraits)}
\label{S:sec:strata}
Main-paper Sec.~5 reports that the cast is significant in every skin-luma stratum and that our
model stays at most $4.0\%$; Table~\ref{S:tab:strata} gives the per-stratum breakdown. Strata are
defined by a skin-luma threshold (dark $<0.45$, medium $0.45$--$0.62$, light $>0.62$); as disclosed
in main-paper Sec.~7 this is a luma proxy, not a colorimetric (ITA) or phototype (Fitzpatrick)
label.

\begin{table}[t]
\centering
\caption{Per-stratum degraded-input audit over 70{,}000 FFHQ portraits. Green rate with Wilson
95\% interval and mean skin $\Delta a^*$. The severity ordering (light $>$ medium $>$ dark)
matches the 70-portrait protocol; our loss reduces every stratum to $\le 4.0\%$.}
\label{S:tab:strata}
\setlength{\tabcolsep}{6pt}
\begin{adjustbox}{max width=\textwidth,center}%
\begin{tabular}{lrcccc}
\toprule
 & & \multicolumn{2}{c}{baseline (degraded)} & \multicolumn{2}{c}{ours (degraded)} \\
\cmidrule(lr){3-4}\cmidrule(lr){5-6}
stratum & $n$ & green rate [Wilson95] & $\Delta a^*$ & green rate [Wilson95] & $\Delta a^*$ \\
\midrule
dark   & 10{,}404 & 56.2\% [55.3, 57.2] & $-4.92$ & 2.8\% [2.5, 3.1] & $+1.81$ \\
medium & 40{,}895 & 64.1\% [63.6, 64.5] & $-6.16$ & 1.2\% [1.1, 1.3] & $+1.41$ \\
light  & 18{,}701 & 83.7\% [83.1, 84.2] & $-8.14$ & 4.0\% [3.7, 4.3] & $-1.10$ \\
\bottomrule
\end{tabular}%
\end{adjustbox}
\end{table}

\noindent\textbf{Clean-input trade-off (disclosed).} On clean inputs the dark stratum is
near-neutral for the baseline ($+0.86$, 18.2\%) and shows the largest magenta shift for our
model ($+5.96$). This is the per-stratum manifestation of the neutral-regime limitation disclosed in
main-paper Sec.~7: on inputs that carry little or no green, the correction pushes nearly neutral
skin toward magenta. Dark skin ends most magenta because it starts near-neutral; the push itself is
smallest there ($+5.10$ vs.\ $+5.70$ medium and $+6.52$ light).

\section{Loss Generalization Across Three Architectures}
\label{S:sec:arch}
Main-paper Sec.~5 states that the corrective loss is not specific to the eLIR family and points
here for the per-architecture table. We compare three ultra-light architectures with distinct
conditioning mechanisms. Each forms a baseline/{+loss} pair that shares the same data and base
reconstruction objective, so that within a pair only the added skin green-excess loss
($\mathcal{L}_{\text{skin}}$, $\tau{=}5$) is varied. CSRNet and NeurOp train both members from
scratch under the synthetic self-supervision protocol; for eLIR (v1g) the pair is our main-paper
model, whose $+$loss member resumes from the baseline rather than training from scratch. NeurOp is
the Neural Color Operators architecture used at its native $28.1$K-parameter size (the original is
already $\sim$28K; we do not reduce it) and CSRNet is the published $36.5$K-parameter model; only
the synthetic recipe and the loss setting are ours.
Table~\ref{S:tab:arch} shows that all three develop the green cast under the synthetic recipe and
that the same loss suppresses it to near-neutral on both evaluation sets.

\begin{table}[t]
\centering
\caption{The synthetic recipe induces the cast and the same $\mathcal{L}_{\text{skin}}$ suppresses it,
across three conditioning mechanisms. Format: synthetic baseline $\rightarrow$ +loss. 70-proto
is the 70-portrait GT-anchored protocol (counts /70); CelebA-HQ is the 500 held-out real photos
(rate).}
\label{S:tab:arch}
\setlength{\tabcolsep}{5pt}
\begin{adjustbox}{max width=\textwidth,center}%
\begin{tabular}{llrll}
\toprule
architecture & mechanism & params & 70-proto: green, $\Delta a^*$ & CelebA-HQ: rate, $\Delta a^*$ \\
\midrule
eLIR (v1g) & histogram-conditioned & 6{,}825 & 52/70, $-7.02 \to$ 1/70, $+0.13$ & 64.4\%, $-6.95 \to$ 0.8\%, $+0.63$ \\
CSRNet & CNN-conditioned & 36{,}489 & 23/70, $-2.93 \to$ 2/70, $+1.16$ & 21.4\%, $-2.59 \to$ 0.4\%, $+1.47$ \\
NeurOp & sequential color operator & 28{,}108 & 52/70, $-7.05 \to$ 4/70, $+1.60$ & 52.0\%, $-4.78 \to$ 2.2\%, $+0.49$ \\
\bottomrule
\end{tabular}%
\end{adjustbox}
\end{table}

\noindent Three consistency notes. (i) The clean-input magenta trade-off appears in all three
architectures (CelebA-HQ clean $\Delta a^*$: eLIR $+3.82$, CSRNet $+3.03$, NeurOp $+4.37$),
i.e.\ the behavior is a property of the correction and not of one architecture. (ii) The loss
lowers degraded-input PSNR (eLIR $17.39\to16.90$, CSRNet $18.85\to17.77$, NeurOp $19.79\to17.72$);
validation PSNR is maintained for CSRNet ($25.13\to25.23$) but drops for NeurOp ($28.09\to26.65$).
(iii) On clean inputs the synthetic baselines differ in sign: eLIR (v1g) and NeurOp shift skin
toward green ($-1.83$, 39/70 flagged; $-1.82$, 41/70), whereas CSRNet is magenta on average
($+1.63$; 13/70 flagged). The NeurOp entry
above is trained on synthetic self-supervision only and is therefore used solely as evidence that
a third mechanism develops the cast and that the loss generalizes; it is not part of the
two-architecture, two-recipe localization experiment of main-paper Table~2.

\paragraph{Real-pair stress test (NeurOp).} Outside the observational $2{\times}2$ (which uses eLIR and
CSRNet), we additionally trained NeurOp on real FiveK pairs. Unlike eLIR and CSRNet ($-0.96$, $+0.03$),
the real-pair NeurOp remains green in-domain on FiveK (skin $\Delta a^*$ $-3.00$, $142/490$ flags; the
synthetic NeurOp scores $-7.05$ on the 70-portrait protocol, a different evaluation set); adding the
loss to this still-green model overshoots to magenta ($+7.96$, PSNR $22.50\to21.81$), a stronger
instance of the magenta over-correction disclosed in main-paper Sec.~7. Full neutralization and the
loss's calibration are therefore architecture-dependent, which is why the cross-architecture results are treated as
observational and the controlled training-regime conclusion is restricted to the matched FiveK/eLIR
setting of Sec.~\ref{S:sec:regime}.

\section{Failed Remedies and the Input-Anchored Sweep}
\label{S:sec:failed}
Main-paper Sec.~4.1 notes that enriching the conditioning signal or adding residual modulation
did not remove the cast, and that a sweep over input-anchored color-preservation weights is
reported here. Table~\ref{S:tab:failed} details the architecture and ensemble variants. These use
an earlier 40-portrait GT-anchored diagnostic protocol and its counts are \emph{not} directly
comparable to the 70-portrait counts elsewhere; the variant names match Fig.~4 of the main
paper.

\begin{table}[t]
\centering
\caption{Architecture / ensemble variants under the 40-portrait GT-anchored audit. Enriching or
replacing the conditioning signal, adding residual modulation, or recombining a
luminance/chrominance ensemble does not remove the cast.}
\label{S:tab:failed}
\setlength{\tabcolsep}{6pt}
\begin{adjustbox}{max width=\textwidth,center}%
\begin{tabular}{llccl}
\toprule
variant & change & green/40 & PSNR & verdict \\
\midrule
v1g-16 (baseline) & --- & 29 & 17.77 & green \\
v1g-22 & 22-bin histogram & 34 & 16.0--18.2 & green (worse) \\
residual & residual modulation & 29 & 18.12 & green (best PSNR, cast intact) \\
YCbCr-hist & YCbCr-histogram conditioning & 34 & --- & green (worse) \\
dual-hist & dual RGB+YCbCr histograms & 34 & --- & green (worse) \\
Y/C ensemble & original-Y + v1g CbCr & 29 & 16.76 & green identical, PSNR worse \\
\bottomrule
\end{tabular}%
\end{adjustbox}
\end{table}

\paragraph{Input-anchored color-preservation sweep.}
An alternative to a corrective loss is to anchor the output hue to the input. We swept the
hue-preservation weight $\tau_{\mathrm{hue}}$: at $\tau_{\mathrm{hue}}{=}0.5$ the green flag is
removed but restoration PSNR drops by about $1.1$\,dB and skin brightness is visibly affected;
at $\tau_{\mathrm{hue}}{\in}[0.2,0.3]$ the cast is only partly resolved and training is
unstable; and re-tuning for quality reintroduces the skin-brightness side effect. This is the
evidence for the main-paper claim that anchoring the output to the (degraded) input directly
opposes the very correction the network must perform. This sweep was run on an earlier
small-sample benchmark; its numbers should not be mixed with the 40- or 70-portrait audit
counts.

\section{Out-of-Distribution Synthetic Probe}
\label{S:sec:ood}
As a controlled negative check we evaluate on 480 synthetic portraits from the public PFHQ
dataset\footnote{PFHQ: privacy-preserving synthetic faces generated with ControlNet-conditioned
Stable Diffusion, from the BFRffusion blind-face-restoration work (arXiv:2312.15736).} (verified
disjoint from the training corpus by perceptual hashing; this is a disjoint synthetic set, not a
held-out real-photo set). The degraded-input cast is largely absent for the baseline
($13.1\%$ flags, mean skin $\Delta a^*$ $-0.74$) because the baseline is already magenta on
clean PFHQ inputs ($+3.08$), so the degradation's shift ($-3.8$) nets out near zero; we draw no
distributional conclusion from this synthetic set. Our corrected model exhibits the neutral-regime behavior
disclosed in main-paper Sec.~7, trading residual flags ($63\to0$) for a magenta shift
($\Delta a^*$ $+6.44$) --- the same direction as in the FiveK ($+4.48$) and clean-input
($+4.4$) regimes. Signed $\Delta a^*$ is again the necessary companion to flag counts.

\section{Statistical Robustness}
\label{S:sec:stats}
\begin{figure}[t]
\centering
\includegraphics[width=\linewidth]{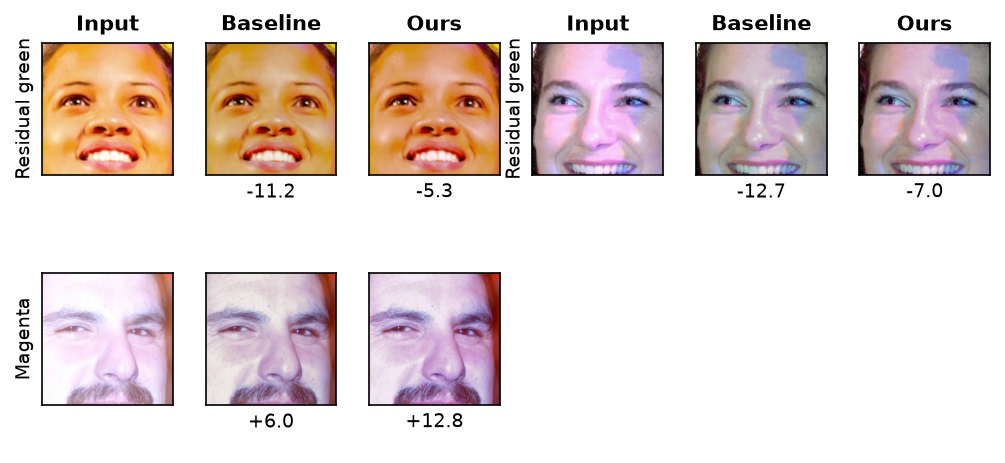}
\caption{FFHQ examples of residual behavior after correction: two residual green cases and a
magenta over-correction (main-paper Sec.~7), chosen to show these behaviors. The candidate lists
were ranked in advance (residual green by final whole-frame green\_shift, magenta by final skin
$\Delta a^*$); going down them we skipped a face not cleared by our age screening, one excluded as a
precaution on re-inspection, and two whose final skin $\Delta a^*$ was near zero ($+0.6$, $-0.7$).
Later magenta candidates were not age-screened, so one magenta example is shown. Numbers: mean
skin $\Delta a^*$.}
\label{S:fig:residual}
\end{figure}

Rates are reported with Wilson 95\% intervals where intervals are given; mean skin $\Delta a^*$ carries
bootstrap 95\% intervals in the primary held-out evaluation (Table~\ref{S:tab:celeba}), while the
stratified and generalization tables report $\Delta a^*$ point estimates. For the
paired 70-portrait protocol we additionally run an exact McNemar test on the green flags. The
paired contingency is: both models flagged $=1$, baseline-only $=51$, ours-only $=0$, neither
$=18$; i.e.\ our model clears 51 of the 52 baseline flags and introduces \emph{no} new ones
(the single remaining flag was already present in the baseline). The exact two-sided McNemar
$p < 10^{-15}$ ($p = 8.9\times10^{-16}$). Figure~\ref{S:fig:residual} shows FFHQ examples of the
two residual behaviors --- residual green and magenta over-correction.

\section{Same-Corpus Training-Regime Control (FiveK)}
\label{S:sec:regime}
Main-paper Sec.~5 compares training recipes with a $2\times2$
architecture-by-regime design whose two regimes also differ in training corpus. This section
removes that confound. Holding the corpus (MIT-Adobe FiveK, 4{,}500 Expert-C targets), dataset
size, architecture (eLIR), objective, schedule (40K iterations, batch 16, learning rate
$5\times10^{-4}$), and seeds fixed, only the \emph{input generation} differs between two cells:
real raw captures paired with their expert targets (cell $R$) versus our synthetic degradation of
the same targets (cell $S$). We trained three seeds per cell and cross-evaluated every model on
identical held-out FiveK test inputs (500 images; $N{=}488$ with a valid ground-truth skin mask,
fixed before evaluation) under three input types: the real raw capture (\emph{raw}, cell~$R$'s
native input), the synthetic degradation of the target (\emph{syndeg}, cell~$S$'s native input),
and the undegraded target (\emph{clean}). Skin $\Delta a^*$ is measured against the expert
target; where syndeg metrics are averaged over five degradation draws per image, the average is
taken before an image-level bootstrap.

\begin{table}[t]
\centering
\caption{Same-corpus control: mean skin $\Delta a^*$ (negative is green) of each training cell on
each shared input, seeds 0/1/2; PSNR (dB, seed 0) in parentheses. Bold marks each cell's native
input. The real-pair cell is near-neutral in its own domain; the synthetic cell is consistently
greener than the real-pair cell on every shared input.}
\label{S:tab:regime_cells}
\setlength{\tabcolsep}{6pt}
\begin{adjustbox}{max width=\textwidth,center}%
\begin{tabular}{lccc}
\toprule
training cell & raw input & syndeg input & clean input \\
\midrule
$R$ (real pairs) & $\mathbf{-0.99/{-}0.78/{-}0.83}$ (22.4) & $+2.39/{+}3.84/{+}2.15$ (12.9) & $+6.64/{+}7.81/{+}5.12$ (13.3) \\
$S$ (synthetic) & $-4.73/{-}4.74/{-}4.36$ (14.0) & $\mathbf{-1.33/{-}0.91/{-}0.90}$ (24.1) & $+0.42/{+}1.53/{+}1.76$ (29.1) \\
\bottomrule
\end{tabular}%
\end{adjustbox}
\end{table}

\paragraph{Training-regime effect.}
Because both cells are evaluated on identical inputs, the paired per-image difference $(S{-}R)$
isolates the training regime. Table~\ref{S:tab:regime_effect} reports it for both fixed input types
together with its factorial decomposition: the training-regime effect (the mean of the raw-input
and syndeg-input gaps), the train-by-input interaction (their difference), and the input-domain
effect. All six shared-input gaps and all three training-regime effects exclude zero
(image-level bootstrap 95\% CIs), whereas every full-set interaction includes zero: the synthetic
regime produces a consistently greener mapping (about $-3.7$ $\Delta a^*$ units) whichever input
both cells receive, so an out-of-distribution failure of one cell alone is an unlikely
explanation. The native-input diagonal ($R$ on raw vs.\ $S$ on syndeg) is near-neutral on the
full set because the training effect ($-3.8$) and the input-domain effect ($+3.6$) oppose each
other; the diagonal is therefore not an estimand of the training effect.

\begin{table}[t]
\centering
\caption{Training-regime effect on skin $\Delta a^*$ (negative $=$ synthetic cell greener), mean
[bootstrap 95\% CI], $N{=}488$ images. Five-draw rows average the syndeg metric over five
degradation draws per image. Facial-skin rows restrict the mask to detected face boxes
($N{=}78$); the masks are computed once on the common target and applied identically to every
model, seed, input type, and draw.}
\label{S:tab:regime_effect}
\setlength{\tabcolsep}{5pt}
\begin{adjustbox}{max width=\textwidth,center}%
\begin{tabular}{lccc}
\toprule
quantity & seed 0 & seed 1 & seed 2 \\
\midrule
$(S{-}R)$ on raw input & $-3.74$ [$-4.09$, $-3.39$] & $-3.96$ [$-4.34$, $-3.58$] & $-3.53$ [$-3.90$, $-3.17$] \\
$(S{-}R)$ on syndeg input & $-3.72$ [$-4.45$, $-3.01$] & $-4.75$ [$-5.60$, $-3.91$] & $-3.05$ [$-3.96$, $-2.16$] \\
\midrule
training-regime effect & $-3.73$ [$-4.18$, $-3.28$] & $-4.36$ [$-4.84$, $-3.86$] & $-3.29$ [$-3.80$, $-2.80$] \\
train$\times$input interaction & $+0.02$ [$-0.68$, $+0.71$] & $-0.79$ [$-1.65$, $+0.08$] & $+0.48$ [$-0.45$, $+1.40$] \\
input-domain effect & $+3.38$ & $+4.23$ & $+3.22$ \\
\midrule
training-regime effect, 5 draws & $-3.62$ [$-3.96$, $-3.28$] & $-4.18$ [$-4.54$, $-3.81$] & $-3.15$ [$-3.50$, $-2.81$] \\
facial-skin subset, 5 draws & $-5.44$ [$-6.46$, $-4.50$] & $-5.84$ [$-6.87$, $-4.85$] & $-4.35$ [$-5.29$, $-3.47$] \\
facial-skin native-input diagonal & $-2.39$ [$-3.82$, $-0.92$] & $-1.98$ [$-3.45$, $-0.61$] & $-1.50$ [$-2.75$, $-0.31$] \\
\bottomrule
\end{tabular}%
\end{adjustbox}
\end{table}

\paragraph{Facial skin.}
FiveK is not face-dominated, so we also restricted the mask to detected face boxes (Haar cascade
run once on the common target; boxes shrunk by 15\% per side; mask $=$ box $\cap$ YCbCr rule;
$N{=}78$ images). Following a preliminary analysis ($N{=}71$) whose face detection did not match
the evaluation preprocessing, we re-ran detection on the evaluation-preprocessed common target
and required at least 256 box-$\cap$-YCbCr skin pixels per image; all facial results reported here
use this revised procedure. The training-regime effect is amplified within facial skin, and there the
native-input diagonal is also significantly negative in all three seeds, i.e.\ the synthetic cell
renders facial skin greener even when each cell is evaluated on its own input. Facial-subset
interactions were seed-dependent ($+0.20/{+}1.10/{+}3.56$; seed 2 significant), so we make no
input-invariance claim for that subset.

\paragraph{External replication and the corpus hypothesis.}
Evaluated zero-shot on the 70-portrait protocol of main-paper Sec.~3.2, the synthetic cell
reproduced the portrait cast in all three seeds (65/63/53 of 70 flagged; skin $\Delta a^*$
$-8.92/{-}8.46/{-}7.48$; the 155K-corpus model of main-paper Table~1: 52/70, $-7.02$). The
real-pair cell's zero-shot outputs on this protocol are far out of domain (PSNR
$\approx 9$--$10$\,dB) and are not interpreted; its neutrality is claimed in-domain only
(Table~\ref{S:tab:regime_cells}). Because FFHQ is absent from this corpus and FiveK is not
face-dominated, FFHQ-specific face-heavy training content is not a necessary explanation of the
cast, although we do not claim it is irrelevant.

\paragraph{A PPR10K-derived control reverses the sign.}
As an exploratory single-seed replication on a different corpus we repeated the design on
1{,}320 PPR10K-derived before/after portrait pairs (source-grouped split fixed in advance;
295 valid test images), with cell $R'$ mapping the raw capture to the retouched target and cell
$S'$ mapping a synthetic degradation of the target to the target. The synthetic cell again
reproduced the external portrait cast (68/70, skin $\Delta a^*$ $-8.97$), but the relative
training-regime effect reversed: $S'{-}R'=+2.25$ [$+1.92$, $+2.58$], with $R'$ itself green on
its native input ($-2.05$). The retouched targets of this derivative set shift skin redward along
$a^*$ relative to their inputs ($+1.41$ [$+1.04$, $+1.76$]), so the real-pair cell must learn a
reddening map whose undershoot appears as a green residual; this suggests a corpus/target-style
interaction whose precise source remains unresolved. We therefore restrict the causal conclusion
to the matched FiveK setting and make no claim of universality across target styles.

\section{Regional and Symmetric Loss Ablations}
\label{S:sec:lossabl}
Main-paper Table~1 compares our loss with a global-mean green loss. Here we test whether the
\emph{asymmetry} of the skin term is what suppresses the cast, by comparing it under matched
conditions against symmetric and regional alternatives, including the closest published prior
(PPR10K's human-region-weighted objective). Every row resumes from the same baseline checkpoint
(52/70, $-7.02$) and trains for a further 150K iterations with identical data, schedule, and
learning rate, exactly as in the main paper; a no-auxiliary continuation controls for the extra
training itself. Coefficients are \emph{initial-gradient-calibrated}: at the shared initialization
we measured each term's gradient norm over 100 fixed batches and set its coefficient to match our
term at $\tau{=}5$ (symmetric green 8.32, run at 8; PPR10K-form 5.70; Lab-$a^*$ 0.05). Robustness to the
calibration is checked by also training the $0.5\times$ and $2\times$ neighbours of each
coefficient and comparing all three at matched 75K checkpoints (Table~\ref{S:tab:lossabl_pm}). The
symmetric green objective is our loss with $\mathrm{ReLU}(\cdot)$ replaced by $|\cdot|$ (same
mask, residual, and normalization; verified numerically). ``PPR10K-form'' reproduces the
geometry of the official PPR10K implementation, $\mathrm{mse}(W\hat{y},Wy)$, i.e.\ a $W^2e^2$
weighting averaged over all pixels, with our skin mask in place of the human-region mask; a
normalized variant ($We^2/\sum W$) and an HRP-inspired mask-matched $L_1$ term are reported for
completeness.

\begin{table}[t]
\centering
\caption{Loss ablation on the 70-portrait protocol. All rows below the resume point continue from the
same baseline for 150K further iterations; ``c'' marks an initial-gradient-calibrated coefficient. Deg./clean
$=$ mean skin $\Delta a^*$ on degraded/clean inputs; CIEDE2000 and PSNR on the degraded pass.
Regional weighting and symmetric axis-specific fidelity improve general fidelity yet leave the
signed bias; only the one-sided term approaches neutrality.}
\label{S:tab:lossabl}
\setlength{\tabcolsep}{5pt}
\begin{adjustbox}{max width=\textwidth,center}%
\begin{tabular}{lccccc}
\toprule
objective [coeff.] & green/70 & deg.\ $\Delta a^*$ & clean $\Delta a^*$ & skin CIEDE2000 & PSNR \\
\midrule
baseline checkpoint (resume point) & 52 & $-7.02$ & $-1.83$ & 13.17 & 17.39 \\
no auxiliary term, $+$150K & 52 & $-7.05$ & $-1.65$ & 12.98 & 17.59 \\
PPR10K-form skin-weighted MSE [5.7c] & 54 & $-6.41$ & $-1.56$ & 12.04 & 18.24 \\
PPR10K-inspired normalized MSE [55c] & 52 & $-6.81$ & $-1.09$ & --- & 17.49 \\
HRP-inspired mask-matched $L_1$ [5] & 54 & $-6.48$ & $-1.31$ & --- & 18.04 \\
symmetric green $|\cdot|$ [5] & 56 & $-7.27$ & $-2.13$ & --- & 17.38 \\
symmetric green $|\cdot|$ [8c] & 59 & $-7.78$ & $-2.04$ & --- & 17.47 \\
symmetric Lab $|\Delta a^*|$ [0.05c] & 39 & $-5.99$ & $-1.68$ & 11.99 & 17.69 \\
\midrule
ours, one-sided [$\tau{=}1$] & 17 & $-3.40$ & $+1.36$ & --- & --- \\
ours, one-sided [$\tau{=}2$] & 8 & $-1.56$ & $+2.72$ & --- & --- \\
\textbf{ours, one-sided [$\tau{=}5$]} & \textbf{1} & $\mathbf{+0.13}$ & $+4.16$ & 12.91 & 16.90 \\
\bottomrule
\end{tabular}%
\end{adjustbox}
\end{table}

\begin{table}[t]
\centering
\caption{Coefficient-neighbourhood check at matched 75K checkpoints: green/70 (mean skin
$\Delta a^*$). Conclusions are unchanged across a $4\times$ range of each coefficient.}
\label{S:tab:lossabl_pm}
\setlength{\tabcolsep}{6pt}
\begin{adjustbox}{max width=\textwidth,center}%
\begin{tabular}{lccc}
\toprule
objective [$0.5\times$ / calibrated / $2\times$] & $0.5\times$ & calibrated & $2\times$ \\
\midrule
PPR10K-form skin-weighted MSE [2.85 / 5.7 / 11.4] & 53 ($-6.58$) & 54 ($-6.72$) & 53 ($-6.21$) \\
symmetric green $|\cdot|$ [4 / 8 / 16] & 53 ($-6.95$) & 55 ($-7.38$) & 54 ($-7.09$) \\
symmetric Lab $|\Delta a^*|$ [0.025 / 0.05 / 0.10] & 39 ($-6.15$) & 35 ($-5.98$) & 37 ($-6.14$) \\
\bottomrule
\end{tabular}%
\end{adjustbox}
\end{table}

\paragraph{Results.}
Relative to the no-auxiliary control, the PPR10K-form objective improved skin CIEDE2000
($12.98\to12.04$) and PSNR ($17.59\to18.24$), confirming that the term was active, yet left the
signed bias essentially intact (54/70, $-6.41$). Symmetric green-axis objectives did not reduce it
at either coefficient (56 and 59 of 70). The symmetric Lab-$a^*$ objective improved fidelity
further (CIEDE2000 11.99) and partially reduced the bias (39/70, $-5.99$), but a substantial
directional residual remained. Only the one-sided objective approached neutrality: relative to
the no-auxiliary control it shifts skin $\Delta a^*$ by $+7.18$ toward neutral (95\% CI
$[+7.01,+7.35]$) and reduces flags from 52 to 1 (exact McNemar on the paired flags, 51
baseline-only vs.\ 0 ours-only discordant images, $p=8.9\times10^{-16}$). The effect is
dose-dependent and monotonic: $\tau{=}1/2/5$ gives 17/8/1 flags while clean-input
over-correction grows ($+1.4/{+}2.7/{+}4.2$), the suppression--overshoot trade-off discussed in
main-paper Sec.~7. Under the tested settings, masking, regional weighting, and symmetric
axis-specific fidelity were insufficient to remove the cast; we do not claim that asymmetry is
universally necessary.

\section{Threshold Sensitivity of the Green Flag}
\label{S:sec:threshold}
The green flag used throughout (whole-frame green\_shift $>0.015$) was fixed as a diagnostic
before the corrected model was trained; the metric audit of main-paper Sec.~5 subsequently
validated it against skin CIEDE2000. Table~\ref{S:tab:threshold} sweeps the threshold over
$t\in[0.005,0.030]$ on the 70-portrait protocol, the 70{,}000-image FFHQ audit, and the held-out
CelebA-HQ set. Separation between the baseline and our model holds at every threshold on every
set. The underlying distributions are far apart: on the 70{,}000-image audit the median skin
$\Delta a^*$ is $-6.41$ for the baseline (5th--95th percentile $-12.5$ to $-0.86$) versus
$+0.85$ for ours ($-5.12$ to $+6.54$), and the median whole-frame green\_shift is $+0.023$
versus $-0.025$.

\begin{table}[t]
\centering
\caption{Percentage of degraded-input images flagged as green as the flag threshold $t$ varies
(baseline / ours). $t{=}0.015$ is the operating point used in the paper.}
\label{S:tab:threshold}
\setlength{\tabcolsep}{6pt}
\begin{adjustbox}{max width=\textwidth,center}%
\begin{tabular}{lcccccc}
\toprule
evaluation set & $t{=}0.005$ & $0.010$ & $0.015$ & $0.020$ & $0.025$ & $0.030$ \\
\midrule
70 FFHQ portraits & 82.9 / 8.6 & 81.4 / 2.9 & 74.3 / 1.4 & 62.9 / 1.4 & 55.7 / 1.4 & 41.4 / 1.4 \\
FFHQ audit (70{,}000) & 86.2 / 5.1 & 78.3 / 3.3 & 68.1 / 2.2 & 56.9 / 1.5 & 45.5 / 1.0 & 34.7 / 0.7 \\
CelebA-HQ held-out (500) & 83.6 / 2.0 & 76.4 / 1.2 & 64.4 / 0.8 & 51.8 / 0.0 & 42.6 / 0.0 & 32.8 / 0.0 \\
\bottomrule
\end{tabular}%
\end{adjustbox}
\end{table}

\section{Skin vs.\ Non-Skin, Degradation-Jitter, and Capacity Controls}
\label{S:sec:controls}
\paragraph{The bias is concentrated in skin.}
On the 70-portrait protocol (degraded pass) the baseline's signed skin $\Delta a^*$ is $-7.03$,
against $-0.80$ on the non-skin pixels of the same images; the paired skin-minus-non-skin difference
is $-6.23$ [$-7.14$, $-5.32$] (image-level bootstrap 95\% CI, $N{=}70$), so the diagnosed defect is
concentrated in skin rather than being a frame-wide green shift. This concentration reflects where
the degradation lowered chroma and the model under-restores it, not a skin-specific chroma gain (Sec.~\ref{S:sec:mechanism}). After correction, skin moves to
$+0.13$ while non-skin shifts to $+3.74$ (paired difference $-3.62$ [$-4.48$, $-2.74$]): the globally
acting operator can induce a compensatory non-skin shift once skin is corrected, so the correction
itself is not spatially confined to skin. Component-wise absolute residuals on skin (reported
alongside CIEDE2000, not as an additive decomposition of it) move from
$|\Delta L^*|/|\Delta a^*|/|\Delta b^*| = 12.4/7.6/5.1$ to $11.6/5.2/9.0$: the signed $a^*$ bias is
suppressed, the remaining error is dominated by lightness, and $b^*$ worsens, consistent with the
loss addressing the signed green--magenta axis only (main-paper Sec.~7).

\paragraph{White-balance jitter is not necessary.}
The training recipe applies symmetric white-balance jitter among its degradations, a natural
suspect for a chromatic bias. Retraining from scratch with the jitter removed, under an otherwise
identical schedule, still produced the cast (50/70, skin $\Delta a^*$ $-6.57$; matched-schedule
reference with jitter: 54/70, $-7.42$). White-balance jitter is therefore not necessary for the
artifact under the tested recipe; Sec.~\ref{S:sec:mechanism} relates the degraded-input shift to skin
chroma that the model under-restores.

\paragraph{Larger architectures.}
To test whether the artifact is confined to ultra-light capacity, we trained two conventional
retouching/restoration networks from scratch on the same 155{,}792-image corpus with the same
degradation model and the same base loss, without any skin term (seed 0), and evaluated
checkpoints along the trajectory (Table~\ref{S:tab:capacity}). DeepLPF (Moran et al., CVPR 2020;
1.77M parameters, $260\times$ ours) used our exact optimization settings and retained the cast at
convergence with a flag rate matching the 6.8K model (51/70) at reduced magnitude ($-4.29$ vs.\
$-7.02$) and healthy fidelity (PSNR 19.7). MIRNet (Zamir et al., ECCV 2020; 31.8M parameters)
required adapted optimization on a single GPU (patch 128, batch 8, learning rate
$2\times10^{-4}$) and is a stress test rather than an architecture-only control; it too retained
an attenuated cast (22/70, $-2.17$). Neither trajectory is monotonic (both attenuate
mid-training and re-strengthen as the learning rate anneals), so we report the converged 150K
values as primary. Capacity and architecture attenuate the cast but did not eliminate it in these
runs: the artifact is not confined to our 6.8K architecture, while architecture--capacity
interactions remain open.

\begin{table}[t]
\centering
\caption{Larger architectures trained with the same recipe and base loss (no skin term),
70-portrait protocol, degraded pass: green/70, mean skin $\Delta a^*$, PSNR (dB) along the
training trajectory. Reference: eLIR 6.8K baseline 52/70, $-7.02$, 17.39.}
\label{S:tab:capacity}
\setlength{\tabcolsep}{6pt}
\begin{adjustbox}{max width=\textwidth,center}%
\begin{tabular}{lccccc}
\toprule
iteration & 25K & 50K & 75K & 100K & 150K \\
\midrule
DeepLPF, 1.77M & 48, $-5.94$, 14.7 & 30, $-3.84$, 19.4 & 45, $-4.93$, 18.4 & 40, $-3.75$, 19.8 & \textbf{51}, $\mathbf{-4.29}$, 19.7 \\
MIRNet, 31.8M (adapted opt.) & 28, $-2.86$, 18.4 & 2, $+1.93$, 20.7 & --- & 17, $-1.86$, 20.3 & \textbf{22}, $\mathbf{-2.17}$, 20.3 \\
\bottomrule
\end{tabular}%
\end{adjustbox}
\end{table}

\section{Why the Shift Is Green and Appears on Skin}
\label{S:sec:mechanism}
This eval-only analysis was added at camera-ready and revises the review version's description of
the cast as model-injected; main-paper Sec.~3.4 summarizes it, and this section gives the
definitions and the full numbers. All quantities use the frozen baseline and final checkpoints, the
70{,}000-image FFHQ audit ($N{=}69{,}925$ with a valid skin mask), the GT-derived skin mask, and
image-level means of per-pixel differences against the ground truth; no model is trained. The
$69{,}925$ images are the same over which Secs.~\ref{S:sec:celeba}--\ref{S:sec:strata} average skin
$\Delta a^*$; this separate recomputation agrees with those means within $0.02$, and the originally
reported values are kept there. Of the $75$ images without a valid mask, $62$ fall in the dark stratum,
$24$ of them because no skin is detected and their luma defaults to $0$.

\paragraph{Quantities.} For each image we measure skin $\Delta a^*$ of (i) the degraded input itself
($I$, no model), (ii) an exact inverse of the known degradation that leaves clipped values at the
clipping level ($O$; exact wherever no channel clips; on clipped pixels it is a reference, not a bound), and (iii) the baseline and final models. Besides the paper's
degradation (brightness $\times1.45$, gamma $0.65$, saturation $0.82$) we evaluate two variants that
change only the brightness factor, $\times1.30$ (the training maximum) and $\times1.00$ (no clipping).
A model's \emph{chroma gain} is the per-image ratio of mean output to mean input chroma
$C^*_{ab}$ on pixels where no channel clips at the paper's degradation (non-skin: pixels outside the mask with GT
$C^*_{ab}{\ge}15$). Radial and tangential components project each pixel's $\Delta(a^*,b^*)$ onto the
ground-truth hue direction and its $+90^\circ$ rotation (radial${<}0$: chroma loss; tangential${>}0$: hue
turns toward yellow). Skin hue is about $47^\circ$ (mean GT $a^*$ $15.7$, $b^*$ $16.8$).

\begin{table}[t]
\centering
\caption{Skin $\Delta a^*$ on the 70{,}000-image audit (negative $=$ green). Clip: fraction of skin
pixels with a clipped channel. With no clipping ($\times1.00$) the exact inverse is perfect, yet the
baseline nearly keeps the degraded input's shift. The last row is the 70-portrait protocol (this pass
reproduces main-paper Table~1 within $0.02$).}
\label{S:tab:mech}
\setlength{\tabcolsep}{6pt}
\begin{adjustbox}{max width=\textwidth,center}%
\begin{tabular}{lccccc}
\toprule
degradation (brightness) & clip & input $I$ & exact inverse $O$ & baseline & final (ours) \\
\midrule
$\times1.45$ (paper) & 0.54 & $-8.31$ & $-5.78$ & $-6.50$ & $+0.78$ \\
$\times1.30$ & 0.38 & $-7.13$ & $-3.30$ & $-6.23$ & $+0.89$ \\
$\times1.00$ (no clipping) & 0.00 & $-6.48$ & $\phantom{-}0.00$ & $-6.24$ & $-0.09$ \\
clean (undegraded) & --- & --- & --- & $-1.46$ & $+4.35$ \\
\midrule
70 portraits, $\times1.45$ & 0.61 & $-9.28$ & $-6.87$ & $-7.02$ & $+0.11$ \\
\bottomrule
\end{tabular}%
\end{adjustbox}
\end{table}

\begin{table}[t]
\centering
\caption{Per-stratum skin $\Delta a^*$ on the 70{,}000-image audit: degraded input, exact inverse and
baseline at $\times1.45$; input and baseline at $\times1.00$; baseline on clean inputs; and the final
model's shift relative to the baseline at $\times1.45$.}
\label{S:tab:mech_strata}
\setlength{\tabcolsep}{6pt}
\begin{adjustbox}{max width=\textwidth,center}%
\begin{tabular}{lrcccccccc}
\toprule
 & & & \multicolumn{3}{c}{$\times1.45$} & \multicolumn{2}{c}{$\times1.00$} & clean & \\
\cmidrule(lr){4-6}\cmidrule(lr){7-8}\cmidrule(lr){9-9}
stratum & $n$ & clip & input & inverse & baseline & input & baseline & baseline & final$-$baseline \\
\midrule
dark   & 10{,}342 & 0.13 & $-4.32$ & $-0.93$ & $-4.91$ & $-6.02$ & $-4.88$ & $+0.86$ & $+6.71$ \\
medium & 40{,}888 & 0.52 & $-8.08$ & $-5.18$ & $-6.15$ & $-6.68$ & $-6.32$ & $-1.61$ & $+7.55$ \\
light  & 18{,}695 & 0.81 & $-11.02$ & $-9.76$ & $-8.13$ & $-6.31$ & $-6.81$ & $-2.40$ & $+7.02$ \\
\bottomrule
\end{tabular}%
\end{adjustbox}
\end{table}

\paragraph{The model under-restores lost chroma.} Without clipping, the baseline's skin $\Delta a^*$
stays within $0.3$ of its degraded input's ($-6.24$ vs.\ $-6.48$; difference $+0.24$ [$+0.22$, $+0.27$]),
although the exact inverse recovers it completely: its small chroma gain is offset in $a^*$ by a hue
rotation toward yellow (tangential $+0.67\!\to\!+2.13$). Across a saturation sweep at $\times1.45$ the output chroma
follows the input chroma (transfer slope $1.06$ [$1.01$, $1.10$] on the 70-portrait protocol, where
$0$ would mean full inversion and $1$ pass-through); gamma-induced chroma changes are partly undone
(slope $0.70$ [$0.65$, $0.75$]). On clean inputs the baseline's own shift is small ($-1.46$) and is a
hue rotation toward yellow rather than a chroma loss (radial $+0.24$, tangential $+2.42$).

\paragraph{No skin-specific chroma gain.} On unclipped pixels at $\times1.45$ the baseline's chroma gain
on skin and on other chromatic pixels agrees within the pre-specified margin of $\pm0.05$ ($1.136$ vs.\
$1.134$ on the same $67{,}602$ images; paired difference $+0.002$ [$+0.001$, $+0.003$]); re-scoring with
the CelebA-HQ hand-annotated skin masks stays within that margin ($1.111$ vs.\ $1.125$; mask IoU
$0.85$). On the same pixels, however, the degradation lowered mean skin chroma (input ratio $0.82$) but
not that of the compared non-skin chromatic pixels ($1.04$), so a similar gain leaves skin below its
original chroma ($0.93$) and non-skin above it ($1.17$). An under-restored chroma loss lowers both $a^*$
and $b^*$ of skin, and a hue rotation toward yellow (inherited from the degradation at $\times1.45$,
added by the model at $\times1.00$ and on clean inputs) lowers $a^*$ further, so the residual reads as
green. The hue rotation was not compared between skin and non-skin pixels.

\paragraph{Clipping and skin tone.} At $\times1.45$ (Table~\ref{S:tab:mech_strata}), light skin's baseline
shift ($-8.13$) is no larger than the exact inverse's ($-9.76$), whereas dark skin's ($-4.91$) is larger
than both its input's ($-4.32$) and the exact inverse's ($-0.93$). At $\times1.00$ (no clipping, but also
lower exposure) the pattern reverses: the model reduces dark skin's input shift ($-6.02\!\to\!-4.88$) but
adds to light skin's ($-6.31\!\to\!-6.81$), and the baseline's light--dark gap narrows from $3.22$ to
$1.93$. The severity ordering of Sec.~\ref{S:sec:strata} therefore reflects both clipping and the model's
own, condition-dependent behavior (on clean inputs light skin shifts $-2.40$ and dark skin $+0.86$).
This account is exploratory: a pre-specified clipping-only explanation of the ordering was not
supported.

\paragraph{What the loss does.} The skin-masked loss shifts skin $a^*$ by a nearly constant amount per
image: $+7.28$ on the paper's degradation (per-image s.d.\ $0.81$; slope of final on baseline $0.98$,
$r{=}0.97$), $+7.12$ at $\times1.30$ and $+6.15$ without clipping. At $\times1.45$ its mean of $+0.78$ is
the sum of a clipped-pixel contribution ($-1.40$) and an unclipped one ($+2.18$): averaged within each
pixel type, skin stays green where a channel clips ($-2.14$) and turns magenta where none does
($+5.60$); on the 70 portraits the same split gives $-2.48$ and $+6.11$ around a mean of $+0.11$. It restores about a quarter of the
baseline's lost skin chroma along the ground-truth direction (radial component $1.76$ of $7.64$) and
mostly rotates hue toward magenta. A constant shift explains why the degraded-input cast is
suppressed while clean inputs move to magenta (main-paper Sec.~7). Why synthetic self-supervision
yields a model that under-restores chroma and rotates hue remains open.

\section{On-Device Latency}
\label{S:sec:latency}
Table~\ref{S:tab:latency} reports latency of the final 6{,}825-parameter model (35\,KB TFLite) on
a mid-tier mobile NPU. At the 512$^2$ protocol resolution the model runs in $9.2$\,ms with full
delegation (32/32 nodes), and in $67.9$\,ms at 1080p.

\begin{table}[!htb]
\centering
\caption{Inference latency (average) of the 35\,KB model on a mid-tier mobile NPU (MediaTek
Dimensity 8400), TFLite \texttt{benchmark\_model}, warmup 10 / 50 runs (12\,MP CPU: warmup 3 / 10
runs). Histogram computation runs outside the model and is excluded.}
\label{S:tab:latency}
\setlength{\tabcolsep}{10pt}
\begin{adjustbox}{max width=\textwidth,center}%
\begin{tabular}{lcc}
\toprule
resolution & CPU (XNNPACK, 4 threads) & NNAPI $\to$ APU (fp16) \\
\midrule
512$\times$512 (protocol) & 46.3\,ms & \textbf{9.2\,ms} \\
1920$\times$1080 (1080p) & 348.6\,ms & \textbf{67.9\,ms} \\
4000$\times$3000 (full sensor, 12\,MP) & 2{,}035\,ms & OP\_FAILED$^\dagger$ \\
\bottomrule
\end{tabular}%
\end{adjustbox}
\end{table}

\noindent Three caveats. (i) The NNAPI numbers use an fp16-relaxed path;
the numeric-equivalence claim (max error $2.4\times10^{-7}$) is measured on the fp32 CPU path.
(ii) $^\dagger$At 12\,MP the NNAPI path exceeds this APU's activation-memory limit (CPU still
completes in $2.0$\,s); full-resolution photos would use tiling or a downsample--correct--upsample
path, which suffices because the global operator is driven by low-resolution statistics. (iii)
The GPU delegate is unavailable for this model because the histogram branch uses a non-4D concat
that the delegate does not support. These limits do not affect the 512$^2$/1080p measurements.

\section{Additional Qualitative Examples}
\label{S:sec:qual}
\begin{figure}[!htb]
\centering
\includegraphics[width=\linewidth]{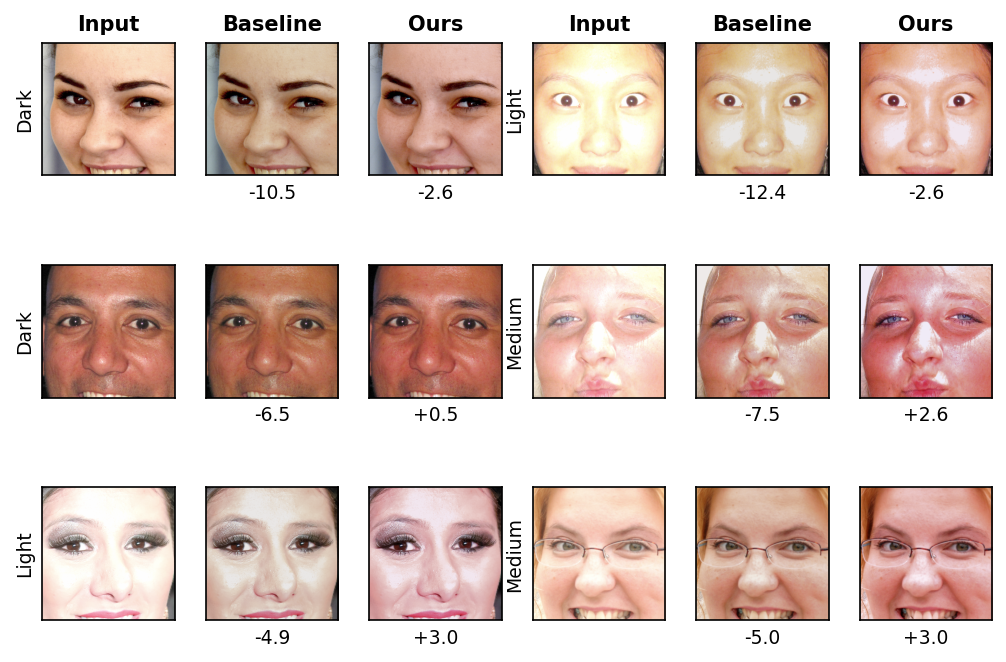}
\caption{Representative corrections across the skin-luma strata and baseline cast severity (mild
to severe), FFHQ portraits. Examples are chosen by a pre-specified score combining baseline green
severity, correction magnitude, a final-magenta penalty, and skin-color fidelity, with skin-tone
and lighting diversity enforced from the top-ranked candidates after manually excluding possible
minors and occluded, stage-lit, tiny or dim-lit faces. Row labels give the ground-truth skin-luma
stratum, which also reflects illumination. The baseline repeatedly leaves skin greener than the
ground truth (not shown); ours moves the signed skin error near neutral. Numbers under each pair: mean
skin $\Delta a^*$ (baseline\,/\,ours).}
\label{S:fig:qual}
\end{figure}

Fig.~\ref{S:fig:qual} shows representative corrections spanning the skin-luma strata and the
mild-to-severe range of baseline cast severity.

\section{Code and Artifact Release}
\label{S:sec:repro}
The code and models will be released through our institution's open-source process, together
with two bundles to support reproduction (checkpoints under non-commercial terms, as they derive
from FFHQ and other research-only datasets): (i)~a reproduction bundle with the skin green-excess loss
($\mathcal{L}_{\text{skin}}$) module, the eLIR-Net (v1g) model with its frozen 6{,}825-parameter
checkpoints (including the 35\,KB TFLite export), and a runnable demo; (ii)~an audit bundle with
per-image metrics, the run commands and environment, and the evaluation scripts used in
Secs.~\ref{S:sec:celeba}--\ref{S:sec:mechanism}. The baseline and final checkpoints are frozen
throughout all evaluations above.

\section{Image Credits}
\label{S:sec:credits}
The faces shown in main-paper Fig.~1 and in Figs.~\ref{S:fig:residual} and~\ref{S:fig:qual} are FFHQ images
derived from the Flickr photographs below; FFHQ aligned and cropped them, and we further cropped
and color-processed the displayed panels. Credits are taken from the FFHQ~v2 metadata of NVIDIA
Corporation (CC BY-NC-SA 4.0); each photograph keeps the license listed here (CC BY 2.0:
\url{https://creativecommons.org/licenses/by/2.0/}; CC BY-NC 2.0:
\url{https://creativecommons.org/licenses/by-nc/2.0/}).

\begin{table}[!htb]
\centering
\caption{Original Flickr photographs of the displayed FFHQ images.}
\label{S:tab:credits}
{\scriptsize
\setlength{\tabcolsep}{3pt}
\begin{tabular}{@{}l p{0.70\linewidth} l@{}}
\toprule
FFHQ ID & title, author, source & license \\
\midrule
20719 & ``Autofix\_IMG\_0672'', Jeesue Kim, \url{https://www.flickr.com/photos/jeesuekim/8274013718/} & CC BY-NC 2.0 \\
36016 & ``RAN Sweetwater Benefit'', Rainforest Action Network, \url{https://www.flickr.com/photos/rainforestactionnetwork/7264896986/} & CC BY-NC 2.0 \\
57991 & ``DSC\_1328'', Tuomas Puikkonen, \url{https://www.flickr.com/photos/darkismus/10666233084/} & CC BY-NC 2.0 \\
40403 & ``Phoenix, AZ'', Ford DSFL, \url{https://www.flickr.com/photos/forddrivingskillsforlife/22945902167/} & CC BY-NC 2.0 \\
16976 & ``Beijing Aug 22-23 086'', Mike Schwarz, \url{https://www.flickr.com/photos/michaelschwarz/4146567062/} & CC BY 2.0 \\
21782 & ``A night out with friends....'', melgupta, \url{https://www.flickr.com/photos/coral/3050788577/} & CC BY-NC 2.0 \\
11340 & ``pic 042'', Gunnar Geir J\'ohannsson, \url{https://www.flickr.com/photos/gunso/4856152475/} & CC BY 2.0 \\
17475 & ``Universidade Federal de Goi\'as'', Cola\c{c}\~oes de Grau UFG, \url{https://www.flickr.com/photos/colacaoufg/40800530425/} & CC BY-NC 2.0 \\
61228 & ``DSC\_0002'', Brian J. McDermott, \url{https://www.flickr.com/photos/22914836@N05/3790951458/} & CC BY 2.0 \\
62895 & ``Family Night 2015'', UrbanPromise, \url{https://www.flickr.com/photos/shannonoberg/22913999540/} & CC BY 2.0 \\
62430 & ``IMG\_9516'', Isidro Urena, \url{https://www.flickr.com/photos/izildur/7203361446/} & CC BY 2.0 \\
28568 & ``Eric \& his manly cowboy hat'', Neil Staeck, \url{https://www.flickr.com/photos/nstaeck/3369414554/} & CC BY-NC 2.0 \\
\bottomrule
\end{tabular}}
\end{table}
\end{document}